\documentclass[acmtog,nonacm]{acmart} %acmart
\acmSubmissionID{xxx}

\usepackage{booktabs} % For formal tables

\usepackage{xcolor} % for color text

\usepackage{wrapfig}
\usepackage[ruled]{algorithm2e} % For algorithms

\SetAlFnt{\small}
\SetAlCapFnt{\small}
\SetAlCapNameFnt{\small}
\SetAlCapHSkip{0pt}

\acmJournal{TOG}
\newcommand{\rev}[1]{#1}
\newcommand{\revv}[1]{#1}

\begin{document}
% Title portion
\title{High-Fidelity 3D Digital Human Head Creation from RGB-D Selfies}

% DO NOT ENTER AUTHOR INFORMATION FOR ANONYMOUS TECHNICAL PAPER SUBMISSIONS TO SIGGRAPH 2019!

\author{Linchao Bao}
\authornotemark[1]
\email{linchaobao@gmail.com}

\author{Xiangkai Lin}
\authornotemark[1]

\author{Yajing Chen}
\authornotemark[1]

\author{Haoxian Zhang}
\authornote{The first four authors contributed equally to this paper.}

\author{Sheng Wang}

\author{Xuefei Zhe}

\author{Di Kang}

\author{Haozhi Huang}

\affiliation{%
 \institution{Tencent AI Lab}
  \city{Shenzhen}
  \country{China}
 }

\author{Xinwei Jiang}
\affiliation{%
 \institution{Tencent NExT Studios}
  \city{Shanghai}
  \country{China}
 }

\author{Jue Wang}

\author{Dong Yu}

\author{Zhengyou Zhang}

\affiliation{%
 \institution{Tencent AI Lab}
  \city{Shenzhen}
  \country{China}
 }

\begin{abstract}
We present a fully automatic system that can produce high-fidelity, photo-realistic 3D digital human heads with a consumer RGB-D selfie camera.
The system only needs the user to take a short selfie RGB-D video while rotating his/her head, and can produce a high quality head reconstruction in less than 30 seconds.
Our main contribution is a new facial geometry modeling and reflectance synthesis procedure that significantly improves the state-of-the-art.
Specifically, given the input video a two-stage frame selection procedure is first employed to select a few high-quality frames for reconstruction.
Then a differentiable renderer based 3D Morphable Model (3DMM) fitting algorithm is applied to recover facial geometries from multiview RGB-D data, which takes advantages of a powerful 3DMM basis constructed with extensive data generation and perturbation.
Our 3DMM has much larger expressive capacities than conventional 3DMM, allowing us to recover more accurate facial geometry using merely linear basis.
For reflectance synthesis, we present a hybrid approach that combines parametric fitting and CNNs to synthesize high-resolution albedo/normal maps with realistic hair/pore/wrinkle details.
Results show that our system can produce faithful 3D digital human faces with extremely realistic details.
The main code and the newly constructed 3DMM basis is publicly available.
\end{abstract}





\begin{teaserfigure}
\centering
\includegraphics[width=1\textwidth]{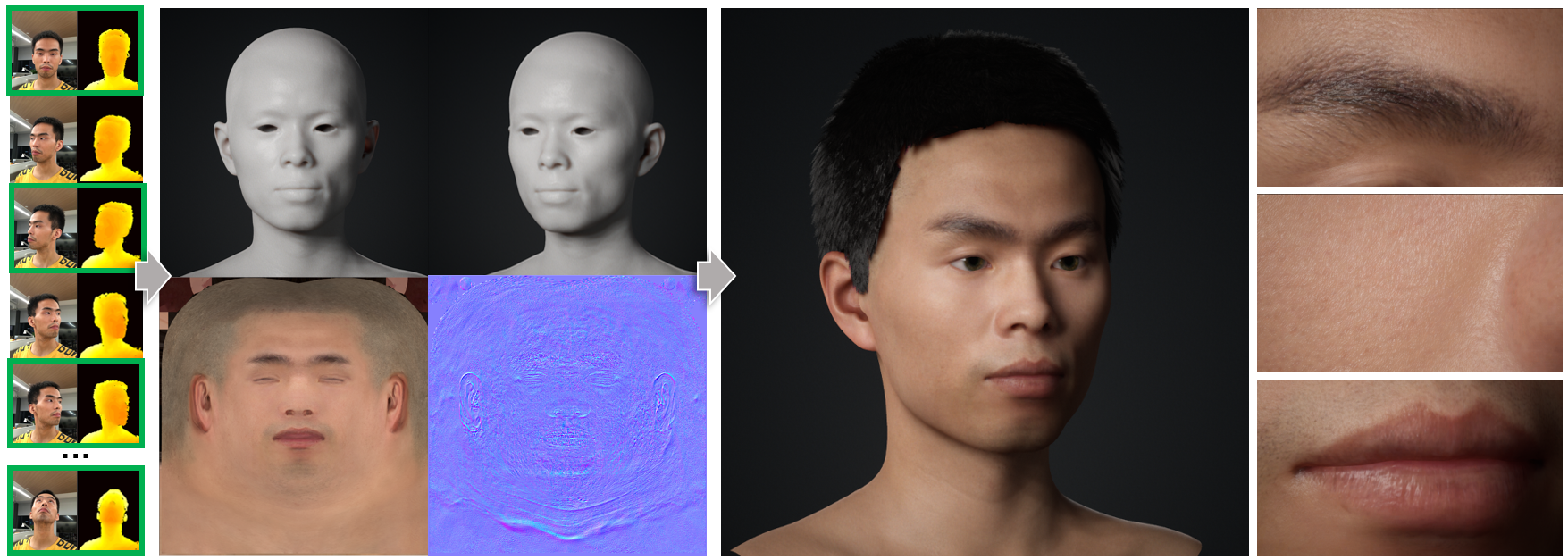}
\vspace{-6mm}
\caption{Our system takes a user's RGB-D selfies as inputs and automatically produce a high-fidelity, riggable head model with high-resolution albedo map and normal map. The model faithfully preserves the user's facial identity features and can be rendered as a realistic digital human character.}
\label{fig:teaser}
\end{teaserfigure}

\maketitle

%%%%%%%%%%%%%%%%%%%%%%%%%%%%%%%%%%
%%%%%%%%%%%%%%%%%%%%%%%%%%%%%%%%%%
\section{Introduction}

Real-time rendering of realistic digital humans is an increasingly important task in various immersive applications like augmented and virtual reality (AR/VR).
To render a realistic human face, high-quality geometry and reflectance data are essential.
There exist specialized hardware like Light Stage \cite{alexander2009digital} for high-fidelity 3D faces capturing and reconstruction in the movie industry, but they are cumbersome to use for consumers.
Research efforts have been dedicated to consumer-friendly solutions, trying to create 3D faces with consumer cameras, e.g., RGB-D data \cite{zollhofer2011automatic,thies2015real}, multiview images \cite{ichim2015dynamic}, or even a single image \cite{hu2017avatar,yamaguchi2018high,lattas2020avatarme}.
While good results have been shown, the reconstructed 3D faces still contain artifacts and are far from satisfactory.

Indeed, faithful 3D facial reconstruction is a challenging problem due to the extreme sensitivity that human perception has towards faces. First, the recovered facial geometry needs to preserve all important facial features like cheek silhouettes and mouth shapes.
Single-image based approaches \cite{hu2017avatar,yamaguchi2018high,lattas2020avatarme} can hardly achieve this due to the lack of reliable geometric constraints.
With multiview RGB/RGB-D inputs, existing approaches \cite{zollhofer2011automatic,thies2015real,ichim2015dynamic} do not fully leverage most recent advances in deep learning and differentiable rendering \cite{genova2018unsupervised,gecer2019ganfit}, leading to inaccurate recovery that does not fully resemble the user's facial shape.
Second, the synthesized facial reflectance maps need to be high-resolution with fine details like eyebrow hair, lip wrinkles, and pore details on facial skin.
Several recent work \cite{saito2017photorealistic,yamaguchi2018high,lattas2020avatarme} have tried to address these issues, but their results still lack natural facial details that are critical for realistic rendering.

In this paper, we present new facial geometry modeling and reflectance synthesis approaches that can produce faithful geometry shapes and high-quality, realistic reflectance maps, from multiview RGB-D data.
Our geometry modeling algorithm extends differentiable renderer based 3DMM fitting, such as GANFIT \cite{gecer2019ganfit}, from single image to multiview RGB-D data.
Different from GANFIT, we employ conventional PCA-based texture bases instead of GAN to reduce the texture space, so that more data constraints can be exerted on geometric shaping.
Additionally, we present an effective frame selection scheme, as well as an initial model fitting procedure, which can avoid enforcing conflicting constraints and increase system robustness.
Moreover, we propose an effective approach that takes advantages of extensive data generation and perturbation to construct the 3DMM, which has much larger expressive capacity compared with previous methods.
We show that even with the linear basis of the new 3DMM, our method can consistently recover accurate, personalized facial geometry.

For facial reflectance modeling, we use high-resolution 2K (2048 $\times$ 2048) UV-maps consisting of an albedo map and a normal map. We propose a hybrid approach that consists of a regional parametric fitting and CNN-based refinement networks.
The regional parametric fitting is based on a set of novel pyramid bases constructed by considering variations in multi-resolution albedo maps, as well as high-resolution normal maps.
Faithful but over-smoothed high-resolution albedo/normal maps can be obtained in this step.
GAN-based networks are then employed to refine the albedo/normal maps to yield the final high-quality results.
Our experiments show that even with the $680 \times 480$ resolution inputs, our method can produce high-resolution albedo/normal maps, where eyebrow hair, lip wrinkles and facial skin pores are all clearly visible.
The high-quality reflectance maps significantly improve the realism of the final renderings in real-time physically based rendering engines.

With the recovered facial geometry and reflectance, we further present a fully automatic pipeline to create a full head rig, by completing a head model, matching a hair model, estimating the position/scale of eyeballs/teeth models, generating the expression blendshapes, etc. We conduct extensive experiments and demonstrate potential applications of our system.

\subsubsection*{Our major contributions include:}
\vspace{-1mm}
\begin{itemize}
  \item A fully automatic system for producing high-fidelity, realistic 3D digital human heads with consumer-level RGB-D selfie cameras. Compared with previous avatar approaches, our system can generate higher quality assets for physically based rendering of photo-realistic 3D characters. The total acquisition and production time for a character is less than 30 seconds. The core code and 3DMM is publicly available\footnote{Code and 3DMM is available at: \url{https://github.com/tencent-ailab/hifi3dface}}.%\footnote{See our project page at: \url{https://tencent-ailab.github.io/hifi3dface_projpage/}}.
  \item A robust procedure consisting of frame selection, initial model fitting, and differentiable renderer based optimization to recover faithful facial geometries from multiview RGB-D data, which can tolerate data inconsistency introduced during user data acquisition.
  \item A novel morphable model construction approach that takes advantages of extensive data generation and perturbation. The constructed linear 3DMM by our approach has much larger expressive capacity than conventional 3DMM.
  \item A novel hybrid approach to synthesize high-resolution facial albedo/normal maps. Our method can produce high-quality results with fine-scale realistic facial details.
\end{itemize}
\vspace{-1mm}

%%%%%%%%%%%%%%%%%%%%%%%%%%%%%%%%%%
%%%%%%%%%%%%%%%%%%%%%%%%%%%%%%%%%%
\section{Related Work}

Creating high-fidelity realistic digital human characters commonly relies on specialized hardware \cite{alexander2009digital,beeler2010high,debevec2000acquiring} and tedious artist labors like model editing and rigging \cite{von2014digital}.
Several recent work seek to create realistic 3D avatars with consumer devices like a smartphone using domain specific reconstruction approaches (i.e., with face shape/appearance priors) \cite{ichim2015dynamic,yamaguchi2018high,lattas2020avatarme}.
We mainly focus on prior arts along this line and briefly summarize the most related work in this section.
Please refer to the recent surveys \cite{egger20193d,zollhofer2018state} for more detailed reviews.

\subsection{Face 3D Morphable Model}

The 3D morphable model (3DMM) is introduced in \cite{blanz1999morphable} to represent a 3D face model by a linear combination of shape and texture bases. These bases are extracted with PCA algorithm on topological aligned 3D face meshes.
To recover a 3D face model from observations, the 3DMM parameters can be estimated instead.
Since the 3DMM bases are linear combinations of source 3D models, the expressive capacity of a 3DMM is rather limited.
Researchers tried to increase the capacity either by automatically generating large amounts of topological aligned face meshes \cite{booth20163d} or turn the linear procedure into a nonlinear one \cite{luthi2017gaussian,Tran_2018_CVPR}.
However, the sampled face models with these 3DMM models are usually flawed \rev{(e.g., distorted mesh surfaces around eyes and mouths due to imperfect alignment \cite{booth20163d}, twisted \cite{luthi2017gaussian} or noisy \cite{Tran_2018_CVPR} face meshes)} and not suitable for realistic digital human rendering.
Another line to increase the expressive capacity of 3DMM is to segment the face into regions and then employ spatially localized bases to model each region  \cite{blanz1999morphable,tena2011interactive,neumann2013sparse}.
We present a novel data augmentation approach that can effectively increase the capacity of either global or localized 3DMM with the same amount of source 3D face meshes as existing approaches.

\subsection{Facial Geometry Capture}

\subsubsection*{Capturing from Single Image}
Given a single face image, the 3D face model can be recovered by estimating the 3DMM parameters with analysis-by-synthesis optimization approaches  \cite{blanz2003face,romdhani2005estimating,garrido2013reconstructing,thies2016face2face,hu2017avatar,garrido2016reconstruction,yamaguchi2018high,gecer2019ganfit}.
A widely adopted approach among them is described in the Face2Face work \cite{thies2016face2face}, where the optimization objective consists of photo consistency, facial landmark alignment, and statistical regularization.
Although there is a recent surge of deep learning based approaches to use CNNs to regress 3DMM parameters \cite{zhu2016face,tran2017regressing,tewari2017mofa,genova2018unsupervised}, the results are commonly not in high fidelity due to lack of reliable geometric constraints.
Some work go beyond the 3DMM parametric estimation to use additional geometric representations to model facial details \cite{richardson2017learning,tran2018extreme,Tewari_2018_CVPR,jackson2017large,sela2017unrestricted,chen2019self,guo2019cnn,kemelmacher20113d,shi2014automatic}, but the results are generally not satisfactory for realistic rendering.

\subsubsection*{Capturing from Multiview Images}
Ichim et al. \shortcite{ichim2015dynamic} present a complete system to produce face rigs by taking hand-held videos with a smartphone. The system relies on a multiview stereo reconstruction of the captured head followed by non-rigid registrations, which is slow and error-prone, especially when motion occurs or no reliable feature points can be detected in face regions.
Recent research on multiview face reconstruction with deep learning methods \cite{dou2018multi,wu2019mvf} do not explicitly model geometric constraints and are not accurate enough for high-fidelity rendering.

\subsubsection*{Capturing from RGB-D Data}
Modeling facial geometries from RGB-D data commonly consists of several separated steps \cite{zollhofer2011automatic,weise2011realtime,bouaziz2013online,li2013realtime,zollhofer2014real}.
First, accumulated point clouds are obtained with rigid registration \cite{newcombe2011kinectfusion}.
Then a non-rigid registration procedure is employed to obtain a deformed mesh from the target mesh model \cite{bouaziz2016modern,chen2013accurate}.
Finally, in order to obtain a 3DMM parametric representation, a morphable model fitting is applied using the deformed mesh as geometric constraints \cite{zollhofer2011automatic,bouaziz2016modern}.
Although the approach is widely adopted as standard practices, it suffers from accumulated errors due to the long pipeline.
Thies et al. \shortcite{thies2015real} propose to use an unified parametric fitting procedure to directly optimize camera poses together with 3DMM parameters, taking into account both RGB and depth constraints.
Their method achieves high-quality results in facial expression tracking, but is not specially designed for recovering personalized geometric characteristics.

\subsection{Facial Reflectance Capture}
Saito et al. \shortcite{saito2017photorealistic} propose to synthesize high-resolution facial albedo maps using CNN style features based optimization like the style transfer algorithm \cite{gatys2016image}.
However, the approach requires iterative optimization and needs several minutes of computation.
Yamaguchi et al. \shortcite{yamaguchi2018high} further propose to inference albedo maps, as well as specular maps and displacement maps, using texture completion CNNs and super-resolution CNNs.
GANFIT \cite{gecer2019ganfit} employs the latent vector of a Generative Adversarial Network (GAN) as the parametric representation of texture maps and then use an differentiable renderer based optimization to estimate the texture parameters.
The most recent work AvatarMe \cite{lattas2020avatarme} propose to infer separated diffuse albedo maps, diffuse normal maps, specular albedo maps, and specular normal maps using a series of CNNs.
We present a novel hybrid approach that can achieve high-quality results while at the same time is more robust than the above pure CNN-based approaches.

\subsection{Full Head Rig Creation}
To complete a full head avatar model, accessories beyond face region need to be attached to the recovered face model, e.g., hair, eyeballs, teeth, etc.
Ichim et al. \shortcite{ichim2015dynamic} describe a simple solution to transfer these accessories (except hair) from a template model and adapt the scales/positions to the reconstructed face model.
Cao et al. \shortcite{cao2016real} use image-based billboards to deal with eyes and teeth, and coarse geometric proxy to deal with hair model. Nagano et al. \shortcite{nagano2018pagan} employ a GAN-based network to synthesis mouth interiors.
Hu et al. \shortcite{hu2017avatar} propose to perform hair digitization by parsing hair attributes from the input image and then retrieving a hair model for further refinement.
There are also some approaches working on modeling hairs in strand level \cite{wei2005modeling,hu2015single,chai2015high,luo2013structure,saito20183d}.
\rev{We in this paper adopt a simplified retrieval approach like Hu et al. \shortcite{hu2017avatar} to attach hair/eyeballs/teeth models.}

For expression blendshape generation, Ichim et al. \shortcite{ichim2015dynamic} present a dynamic modeling process to produce personalized blendshapes, while Hu et al. \shortcite{hu2017avatar} adopt a simplified solution to transfer generic FACS-based blendshapes to the target model.
The expression blendshapes can also be generated with a bilinear 3DMM model like FaceWarehouse \cite{cao2014facewarehouse}, where the face identity and expression parameters are in independent dimensions.
\rev{Wang et al. \shortcite{wang2020facial} recently present a global-local multilinear framework to synthesize high-quality facial expression blendshapes.
For simplicity, we adopt a blendshape transfer approach similar to Hu et
al. \shortcite{hu2017avatar} to obtain expression blendshapes for rigging.}

%%%%%%%%%%%%%%%%%%%%%%%%%%%%%%%%%%
%%%%%%%%%%%%%%%%%%%%%%%%%%%%%%%%%%
\section{Overview}

We first introduce the 3D face dataset used in our system.
Then we describe the goal of our system, followed by the user data acquisition process and a summary of the main processing steps.

\begin{figure*}
\centering
\includegraphics[width=\textwidth]{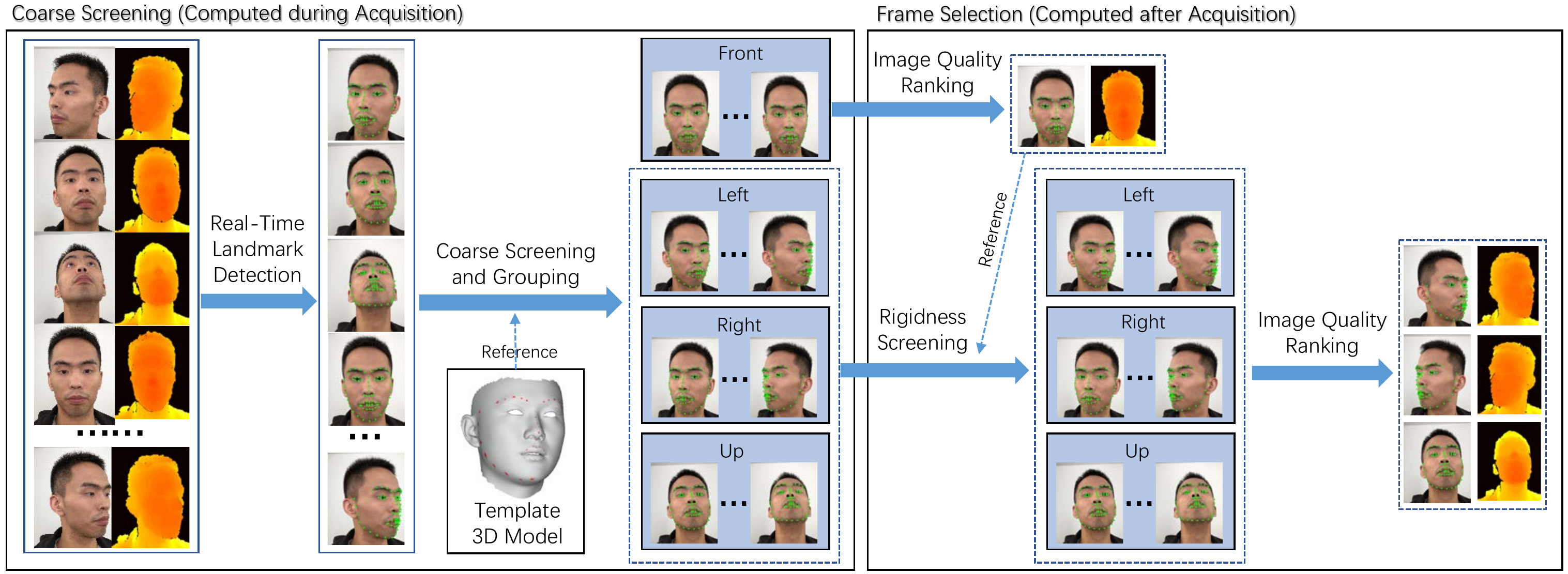}
\vspace{-7mm}
\caption{Our two-stage frame selection procedure. Four frames are selected out of 200-300 frames considering both view coverage and data quality. Note that the reference model used for coarse screening and grouping is a template 3D face model, which may lead to inaccurate pose estimation. But the rough poses are sufficient for excluding extreme/invalid frames and categorizing the rest frames into pose groups. In the second stage, the reference model for rigidness screening is the lifted 3D landmarks from the front face data, which can result more accurate poses for more strict rigidness verification.}
\vspace{-1mm}
\label{fig.frameselectionalgo}
\end{figure*}

\subsubsection*{3D Face Dataset}
We use a specialized camera array system \cite{beeler2010high} to scan 200 East Asians, including 100 males and 100 females, aged from 20 to 50 years old (with their permissions to use their face data).
The scanned face models are manually cleaned and aligned to a triangle mesh template with 20,481 vertices and 40,832 faces.
Each face model is associated with a 2K-resolution (2048 $\times$ 2048) albedo map and a 2K normal map, where pore-level details are preserved.
A linear PCA-based 3DMM \cite{blanz1999morphable} can be constructed from the dataset, which consists of shape basis, albedo map basis, and normal map basis.
Note that we propose a novel approach to construct an augmented version of the 3DMM shape basis in Sec. \ref{sec.3dmmaug}.
Besides, a novel pyramid version of the 3DMM albedo/normal maps is presented in Sec. \ref{sec.textureregionbases}.

\subsubsection*{Goal}
The goal of our system is to capture high-fidelity users facial geometry and reflectance with RGB-D selfie data, which is further used to create and render full-head, realistic digital humans.
For geometry modeling, we use 3DMM parameters to represent a face, since it is more robust to degraded input data and with more controllable mesh quality than deformation-based representations.
For reflectance modeling, we synthesize 2K-resolution albedo and normal maps regardless of the input RGB-D resolution.

\subsubsection*{User Data Acquisition}
We use an iPhone X to capture user selfie RGB-D data.
Note that it is common nowadays for a smartphone to be equipped with a front-facing depth sensor and any such phone can be used.
While a user is taking selfie RGB-D video, our capturing interface will guide the user to consecutively rotate his/her head to left, right, upward, and back to middle.
The entire acquisition process takes less than 10 seconds, and a total of 200-300 frames of RGB-D images are collected, with resolution $640 \times 480$.
The face region for computation is cropped (and resized) to $300 \times 300$.
The camera intrinsic parameters are directly read from the device.

\begin{figure}[H]
\centering
\vspace{-2mm}
\includegraphics[width=0.45\textwidth]{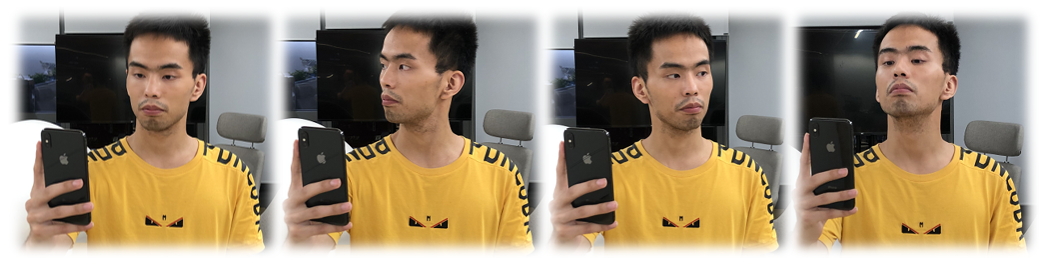}
\vspace{-1mm}
\label{Fig.2}
\end{figure}

\subsubsection*{Processing Pipeline}
We first employ an automatic frame selection algorithm to select a few high-quality frames that cover all sides of the user (Sec. \ref{sec.preprocframeselect}).
Then an initial 3DMM model fitting is computed with the detected facial landmarks in the selected frames (Sec. \ref{sec.initfitting}).
Starting from the initial fitting, a differentiable renderer based optimization with multiview RGB-D constraints (Sec. \ref{sec.optframework}) is applied to solve the 3DMM parameters as well as lighting parameters and poses.
Based on the estimated parameters, high-resolution albedo/normal maps are then synthesized (Sec. \ref{sec.texturesynth}).
Finally, high-quality, realistic full head avatars can be created and rendered (Sec. \ref{sec.fullheadrig}).

%%%%%%%%%%%%%%%%%%%%%%%%%%%%%%%%%%
%%%%%%%%%%%%%%%%%%%%%%%%%%%%%%%%%%
\section{Frame Selection}
\label{sec.preprocframeselect}
% frame selection
% landmark fitting

There are typically 200-300 frames acquired from a user.
For efficiency and robustness, we developed a robust frame selection procedure to select a few high-quality frames for further processing, which considers both view coverage and data quality.
As shown in Fig. \ref{fig.frameselectionalgo}, the procedure consists of two stages as described below.

\subsubsection*{Coarse Screening and Preprocessing}
We first apply a real-time facial landmark detector (a MobileNet \cite{howard2017mobilenets} model trained on 300W-LP dataset \cite{zhu2016face}) on RGB images to detect 2D landmarks for each frame.
Then a rough head pose for each frame can be efficiently computed with the correspondences between the 2D landmarks and the 3D keypoints on a template 3D face model using PnP algorithm \cite{lepetit2009epnp}.
Frames with extreme/invalid poses or closed-eye/opened-mouth expressions can be easily identified and screened out with the 2D landmarks and rough head poses.
We categorize the rest frames by poses into groups: \emph{front}, \emph{left}, \emph{right}, and \emph{up}.
Each group only keeps 10-30 frames near the center pose of the group.
Note that more groups can be obtained by categorizing the frames with finer-level angle partitioning.
We experimented with different number of groups and found four is a good balance between accuracy and efficiency.
The remaining depth images are preprocessed to remove depth values beyond the range between 40cm and 1m (the typical selfie distances).
Bilateral filtering \cite{paris2009fast} with a small spatial and range kernel is then applied to the depth images to attenuate  noises.

\subsubsection*{Frame Selection}
For each group, we further select one frame based on two criteria: image quality and rigidness.
To measure the image quality of a frame, we compute the Laplacian of Gaussian (LoG) filter response and use the variance as a motion blur score (images with a larger score are sharper).
A front face frame is first selected based on the motion blur score in the front group.
We then compute the rigidness between each frame in the other groups and the front face with the help of depth data.
Specifically, the detected 2D landmarks for each frame are lifted from 2D to 3D using depth data.
Note that occluded landmarks are automatically removed according to the group that a frame belongs to, e.g., for a frame in the left group, the landmarks on the right side of the face are removed.
We use RANSAC method to compute the relative pose between each frame in the other groups and the front face using the 3D-3D landmark  correspondences \cite{arun1987least}.
Frames with too many outliers are considered as low rigidness and thus are excluded.
Then a best frame in each group can be found based on the motion blur score.
The output of this step is four frames with the 3D landmarks.

%%%%%%%%%%%%%%%%%%%%%%%%%%%%%%%%%%
%%%%%%%%%%%%%%%%%%%%%%%%%%%%%%%%%%
\section{Facial Geometry Modeling}
\label{sec.geometrymodel}

\begin{figure}[t]
\centering
\includegraphics[width=0.45\textwidth]{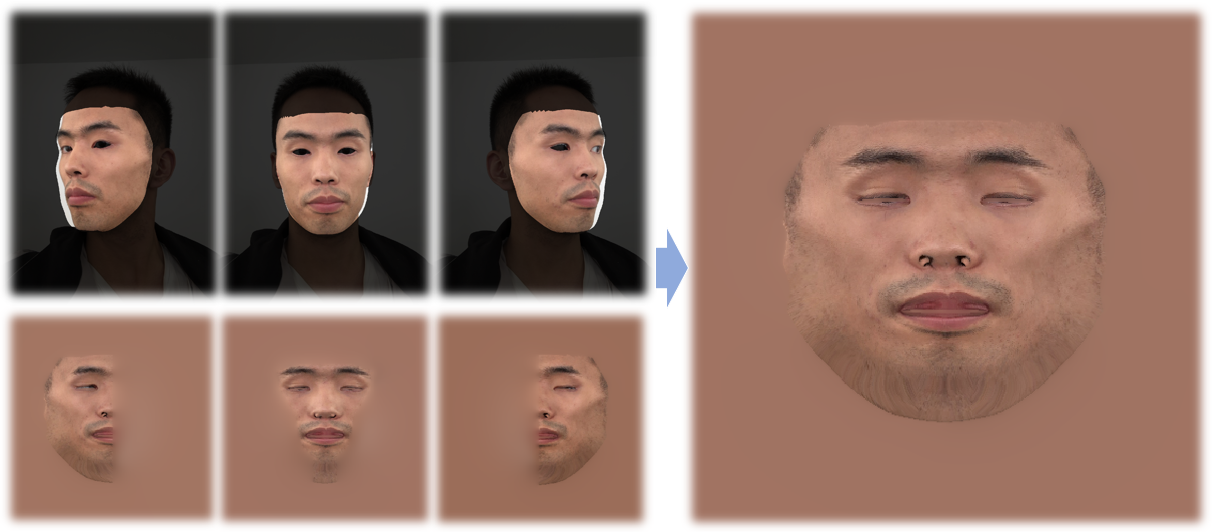}
\vspace{-2mm}
\caption{The masks derived from the detected landmarks for texture blending. We use the verb ``unwrap'' to refer to the process of extracting partial texture maps from input photos and blending them into a complete texture map.}
\label{fig.textmask}
\end{figure}

\subsection{Initial Model Fitting}
\label{sec.initfitting}
We use PCA-based linear 3DMM \cite{blanz1999morphable} for parametric  modeling.
The shape and albedo texture of a face model is represented as
\begin{equation*}\label{eq.3dmm}
  \begin{array}{l}
  \mathbf{s} = \Bar{\mathbf{s}} + S \mathbf{x}_{shp},\\
  \mathbf{a} = \Bar{\mathbf{a}} + A \mathbf{x}_{alb},
  \end{array}
\end{equation*}
where $\Bar{\mathbf{s}}$ is the vector format of the mean 3D face shape model, $S$ is the shape identity basis, $\mathbf{x}_{shp}$ is the corresponding identity parameter vector to be estimated, $\Bar{\mathbf{a}}$ is the vector format of the mean albedo map, $A$ is the albedo map basis, $\mathbf{x}_{alb}$ is the corresponding albedo parameter vector to be estimated. The details of the bases are presented in Secs. \ref{sec.3dmmaug} (shape) and \ref{sec.textureregionbases} (albedo).

We fit an initial shape model with the detected 3D landmarks \rev{(with depth information)} using a ridge regression \cite{zhu2015high}.
A partial texture map can be extracted by projecting the shape model onto each input image.
With a predefined mask derived from landmarks for each view (see Fig. \ref{fig.textmask}), the partial texture maps are then blended into a complete texture map using Laplacian pyramid blending \cite{burt1983multiresolution}.
The initial albedo parameters can be obtained with another ridge regression to fit the blended texture map.

\begin{figure*}
    \centering
    \includegraphics[width=\textwidth]{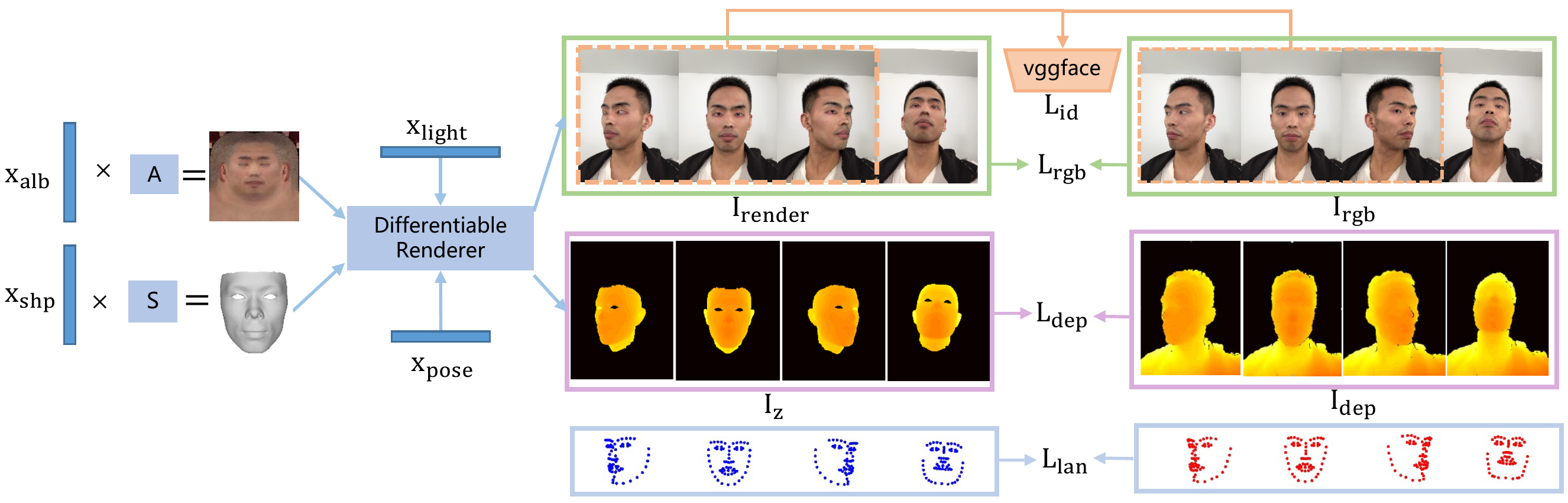}
    \vspace{-6mm}
    \caption{Our optimization framework. The parameters to be solved include: 3DMM parameters $\mathbf{x}_{shp}$ and $\mathbf{x}_{alb}$ for a user, lighting parameters $\mathbf{x}_{light}$ and poses $\mathbf{x}_{pose}$ for each view. The constraints include: landmark loss $L_{lan}$, RGB photo loss $L_{rgb}$, depth loss $L_{dep}$, and identity perceptual loss $L_{id}$.}
    \vspace{-1mm}
    \label{fig:optim}
\end{figure*}

\subsection{Optimization}
\label{sec.optframework}
% optimization framework

Fig. \ref{fig:optim} shows our optimization framework.
The parameters to be optimized are
\begin{equation*}
\mathcal{P} = \{ \mathbf{x}_{shp}, \mathbf{x}_{alb}, \mathbf{x}_{light}, \mathbf{x}_{pose} \},
\end{equation*}
and $\mathbf{x}_{shp} \in \mathbb{R}^{500}$ is the shape parameter, $\mathbf{x}_{alb} \in \mathbb{R}^{199}$ is the albedo parameter, $\mathbf{x}_{light} \in \mathbb{R}^{27}$ is the second-order spherical harmonics lighting parameter, $\mathbf{x}_{pose} \in \mathbb{R}^{6}$ include the rotation and translation parameters for rigid transformation. Note that we have only one $\mathbf{x}_{shp}$ and one $\mathbf{x}_{alb}$ for an user, while the number of $\mathbf{x}_{light}$ and $\mathbf{x}_{pose}$ equals to the number of views.
With a set of estimated parameters and the 3DMM basis, a set of rendered RGB-D frames can be computed via a differentiable renderer \cite{genova2018unsupervised,gecer2019ganfit}.
The distances between the rendered RGB-D frames and the input RGB-D frames can be minimized by backpropagating the errors to update parameters $\mathcal{P}$.
The loss function to be minimized is defined as:
\begin{multline}\label{eq.totaloptloss}
    L(\mathcal{P}) = \omega_{rgb}L_{rgb}(\mathcal{P}) +
    \omega_{dep}L_{dep}(\mathcal{P}) + \\
    \omega_{id}L_{id}(\mathcal{P}) +
    \omega_{lan}L_{lan}(\mathcal{P}) +
    \omega_{reg}L_{reg}(\mathcal{P}),
\end{multline}
where $L_{rgb}(\mathcal{P})$ denotes pixel-wise RGB photometric loss, $L_{dep}(\mathcal{P})$ indicates pixel-wise depth loss, $L_{id}(\mathcal{P})$ is identity perceptual loss,
$L_{lan}(\mathcal{P})$ represents landmark loss, and $L_{reg}(\mathcal{P})$ means regularization terms.
Note that the landmark loss, RGB photometric loss, and regularization term are similar to conventional analysis-by-synthesis optimization approaches \cite{thies2016face2face}.
The identity perceptual loss is also employed in recent differentiable renderer based approaches \cite{genova2018unsupervised,gecer2019ganfit}.
We extend these losses into multiview setting and incorporate depth data for geometric constraints. The details of each term are as follows.

% loss functions

\subsubsection*{RGB Photo Loss} The pixelwise RGB photometric loss is:
\begin{equation*}
    L_{rgb}(\mathcal{P}) = \| I_{rgb} - I_{render}(\mathcal{P}) \|_{2},
\end{equation*}
where $I_{rgb}$ is the input RGB image, $I_{render}$ is the rendered RGB image from the differentiable renderer. We adopt $\ell_{2,1}$-norm because it is more robust against outliers than $\ell_{2}$-norm.

\subsubsection*{Depth Loss} The depth loss is defined as:
\begin{equation*}
    L_{dep}(\mathcal{P}) = \rho(\| I_{dep} - I_{z}(\mathcal{P})) \|_{2}^{2}),
\end{equation*}
where $\rho(\cdot)$ defines a truncated $\ell_{2}$-norm that clips the per-pixel mean squared error, $I_{dep}$ is the input depth image, $I_{z}$ is the rendered depth image from the differentiable renderer.
The truncated function makes the optimization more robust to depth outliers.

\subsubsection*{Identity Perceptual Loss} To capture high-level identity information, we apply identity perceptual loss defined as
\begin{equation*}
    L_{id}(\mathcal{P}) = \| \psi(I_{rgb}) - \psi(I_{render}) \|_{2}^{2},
\end{equation*}
where $\psi(\cdot)$ is the deep identity features exacted from a pretrained face recognition model. Here we use features from the \emph{fc7} layer of VGGFace model \cite{parkhi2015deep}.

\subsubsection*{Landmark Loss} We define the landmark loss as the average distances between the detected 2D landmarks and projected landmarks from the predicted 3D model:
\begin{equation*}
    L_{lan}(\mathcal{P}) = \dfrac{1}{| \mathcal{F} |} \sum_{f_j \in \mathcal{F}} \omega_{j} \|f_j - \Pi(\Phi(v_j))\|_{2}^{2},
\end{equation*}
where $f_j \in \mathcal{F}$ are the detected landmarks, $\Pi(\Phi(v_j))$ denotes that the vertex $v_j$ is rigidly transformed by $\Phi$ and projected by camera $\Pi$. The weighting $\omega_{j}$ is to control the importance of each keypoint, where we set 50 for those located in eye, nose and mouth, while others are 1.

\subsubsection*{Regularization} To ensure the plausibility of the reconstructed faces, we apply regularization to shape and texture parameters:

\begin{equation*}
    L_{reg}(P) = \omega_{shp} \| x_{shp} \|_{2}^{2} + \omega_{alb} \| x_{alb} \|_{2}^{2},
\end{equation*}
where we set $\omega_{shp} = 0.4$ and $\omega_{alb} = 0.001$.

\subsubsection*{Implementation Details}
For efficiency, we use albedo maps of a $512 \times 512$ resolution during the optimization.
We render RGB-D images and compute the pixel losses in the same resolution as input depth images, which is $300 \times 300$.
The weightings in Eq. \eqref{eq.totaloptloss} is set to $\omega_{rgb} = 1000.0$, $\omega_{depth} = 1000.0$, $\omega_{id} = 1.8$, $\omega_{lan} = 10$, $\omega_{reg} = 1.0$.
We use Adam optimizer \cite{kingma2014adam} in Tensorflow to update parameters for 150 iterations to get the results, with a learning rate $0.05$ decaying exponentially in every 10 iterations.

\subsubsection*{Relation to Existing Approaches}
The differences between our approach and state-of-the-art 3DMM fitting approaches are listed in Table \ref{tab:equimethod}.
Result comparisons are presented in Sec. \ref{sec.expresultgeometry}.
Note that our implementation will be publicly available and can be easily configured into equivalent settings to other approaches by changing the combinations of input data and loss terms.

\begin{table}[h]
    \centering
    \footnotesize
    \begin{tabular}{c|c|c|c}
    \hline
        \textbf{Method} & \textbf{Input} & \textbf{Loss Term} &  \textbf{Optimizer}  \\
        \hline
        \hline
        Ours & RGB-D & $L_{rgb},L_{dep},L_{id},L_{lan},L_{reg}$ & DR-based  \\
        \hline
       GANFIT & RGB & $L_{rgb},L_{id},L_{lan},L_{reg}$ & DR-based  \\
        \hline
       Face2Face & RGB & $L_{rgb},L_{lan},L_{reg}$ &  Gauss-Newton \\
        \hline
        \cite{thies2015real} & RGB-D & $L_{rgb},L_{dep},L_{lan},L_{reg}$ & Gauss-Newton \\
        \hline
    \end{tabular}
    \vspace{1mm}
    \caption{Different 3DMM fitting approaches. ``DR-based'' stands for differentiable renderer based optimizer.}
    \vspace{-5mm}
    \label{tab:equimethod}
\end{table}

\subsection{Morphable Model Augmentation}
\label{sec.3dmmaug}
% problem motivation
% algorithm
% evaluation
% mention part-based basis

As the constraints incorporated in the optimization are rich, we found the expressive capacity of the linear 3DMM constructed using conventional approaches are very limited. We here present an augmentation approach to effectively boost the 3DMM capacity.
Our approach is motivated by the observation that human faces are mostly not symmetrical.
This will cause ambiguities when aligning face models.
The reason is that during the alignment of two models, the relative rotation and translation between them is determined by minimizing the errors at some reference points on the models.
Different reference points may lead to different alignment results.
There are no perfect reference points due to the asymmetrical structures of human faces.
This reminds us that we can perturb the relative pose between two aligned models to get an ``alternative'' alignment.
In this way, we can actually get additional samples for PCA, since the new alignments introduces new morphing targets.
Furthermore, we can use a set of perturbation operations including pose perturbation, mirroring, region replacement, etc., to augment the aligned models.
Based on the large amount of generated data, we propose a stochastic iterative algorithm to construct a 3DMM that compresses more capacities into lower dimensions of the basis.

\subsubsection*{Data Generation and Perturbation}
\begin{figure}[t]
\centering
\includegraphics[width=0.4\textwidth]{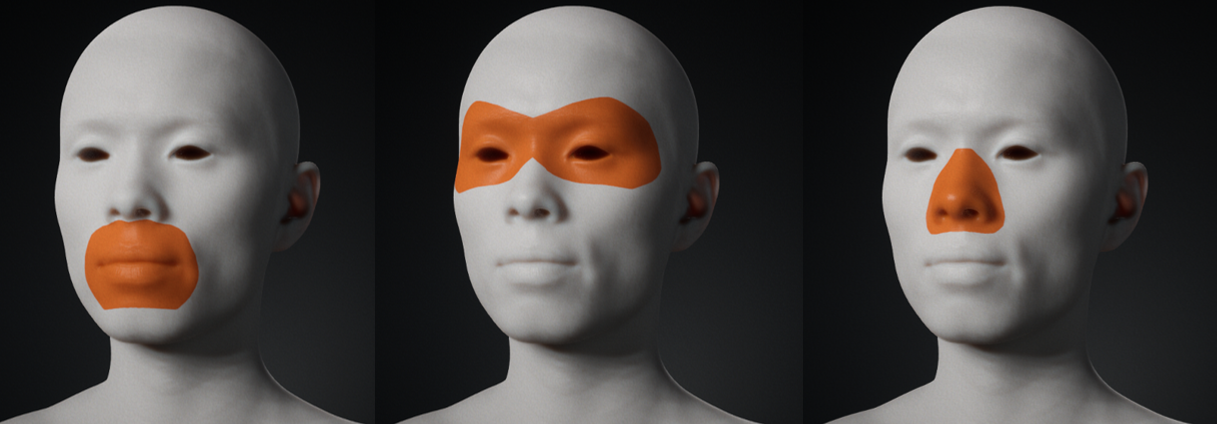}
\vspace{-2mm}
\caption{Masks for region replacement.}
\label{fig.regionmask}
\vspace{-2mm}
\end{figure}

Starting from the 200 aligned face shape models, our data generation and perturbation process consists of the following steps:
\begin{itemize}
  \item \emph{Region Replacement with Perturbation.} We first replace the nose region of each model with other models, with a rotation perturbation along the pitch angle (uniformly sampled within $\pm 1$ degree). Mouth region is also processed in the same way. For eye region, we apply replacement without perturbation. The different perturbations are empirically designed by minimizing the introduced visual defects during processing. The facial regions used in this step are shown in Fig. \ref{fig.regionmask}.
  \item \emph{Rigid Transformation Perturbation.} We then apply rigid transformation perturbations to each face model, where the uniformly sampled range is set to: $\pm 1$ degree along yaw/pitch/roll angles for rotation, $\pm 1\%$ along each of the three axes for translation, $\pm 1\%$ for scale.
  \item \emph{Mirroring.} Finally, we apply a mirroring for all the generated face models along model local coordinate system. In this way, we get over 100,000 face models in total.
\end{itemize}

\subsubsection*{Stochastic Iterative 3DMM Construction}
Our iterative 3DMM construction algorithm is presented in Alg. \ref{alg.iterativealg}.
There are two levels of loops in our algorithm.
We maintain a model set $\mathcal{S}$ for 3DMM construction and update it inside the loops.
In each iteration of the outer loop, we sample a test set $\mathcal{D}$ with $n=1000$ models from the whole generated dataset.
In each iteration of the inner loop, we use the constructed 3DMM from $\mathcal{S}$ to fit models in $\mathcal{D}$, and add $m=25$ models with largest fitting errors in $\mathcal{D}$ into $\mathcal{S}$.
The convergence threshold is empirically set such that the inner loop is usually converged in less than 5 iterations.
Note that in the inner loop, a model sample in $\mathcal{D}$ could be repeatedly added into $\mathcal{S}$ for several times.
In this case, constructing a 3DMM from the final $\mathcal{S}$ is different from directly performing PCA on the whole dataset as the data population is changed.
Our algorithm encourages more data variance to be captured using fewer principal components (note that in each iteration we construct 3DMM using only the principal components with $99.9\%$ cumulative explained variance).

\begin{algorithm}[t]
\SetAlgoLined
 \SetKwInOut{Output}{Output}
 \SetKwInOut{Parameter}{Params}
 \Parameter{$n=1000, m=25$, $Thresh$}
 \Begin{
 Face model set $\mathcal{S}$ $\leftarrow$ initial 200 models\;
 \Repeat{the whole dataset is sampled}{
  Randomly sample (without replacement) a test set $\mathcal{D}$ with $n$ face models from the whole dataset (over 100,000 models)\;
  $k \leftarrow 0$\;
  $\xi \leftarrow \infty$\;
  \While{$\xi > Thresh$}{
   Apply Principal Component Analysis (PCA) on $\mathcal{S}$\;
   Select the principal components with $99.9\%$ cumulative explained variance to get the 3DMM basis $S^k$\;
   Fit the models in $\mathcal{D}$ using basis $S^{k}$\;
   Select the $m$ models with largest fitting errors as set $\mathcal{M}$\;
   Add the the $m$ corresponding mirrored models into $\mathcal{M}$\;
   $\xi \leftarrow $ the median error of the $m$ models\;
   $\mathcal{S} \leftarrow \mathcal{S} \cup \mathcal{M}$\;
   $k \leftarrow k+1$\;
  }
 }
 }
 \Output{PCA basis $S^k$.}
 \caption{Iterative 3DMM Construction Algorithm}
\label{alg.iterativealg}
\end{algorithm}

\begin{figure}[t]
\centering
\includegraphics[width=0.48\textwidth]{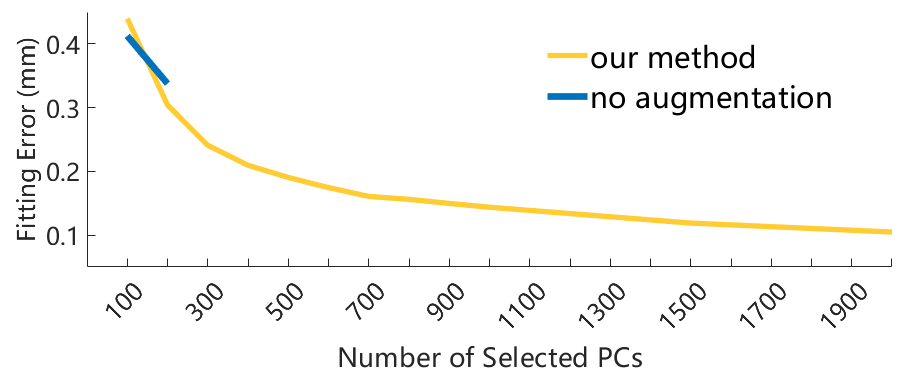}
\vspace{-8mm}
\caption{Mean fitting errors with two versions of basis. Note that the basis without augmentation are constructed from 200 models and thus have a maximum number of dimensions 199. The results show the expressive power of our basis is much larger than original basis.}
\label{fig.pcafittingerror}
\vspace{-2mm}
\end{figure}

\begin{figure}[t]
\centering
\includegraphics[width=0.43\textwidth]{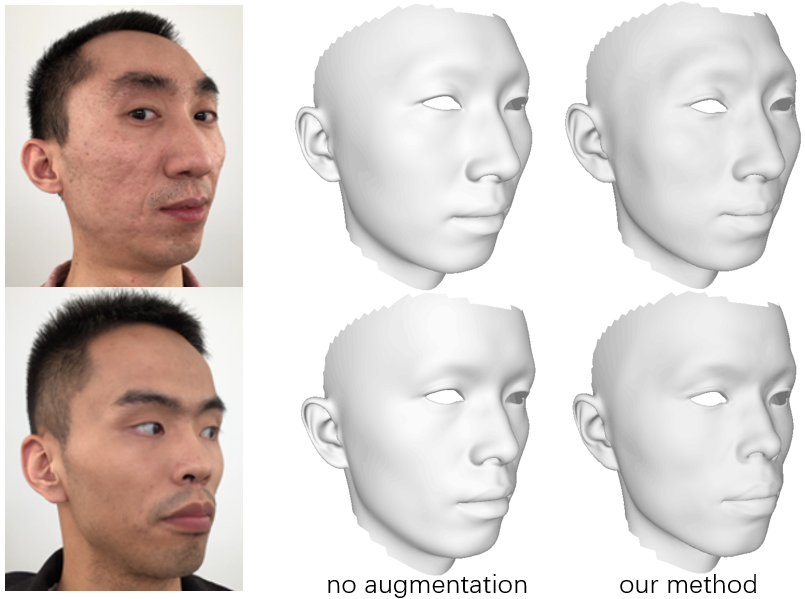}
\vspace{-1mm}
\caption{The recovered geometries with two versions of basis. The basis obtained with our method can preserve more personalized facial geometries (note the regions of facial silhouette, mouth shape, and the nose shape).}
\label{fig.augmentcompare}
\vspace{-2mm}
\end{figure}

\subsubsection*{Evaluation}

\begin{figure*}[t]
    \centering
    \includegraphics[width=0.98\textwidth]{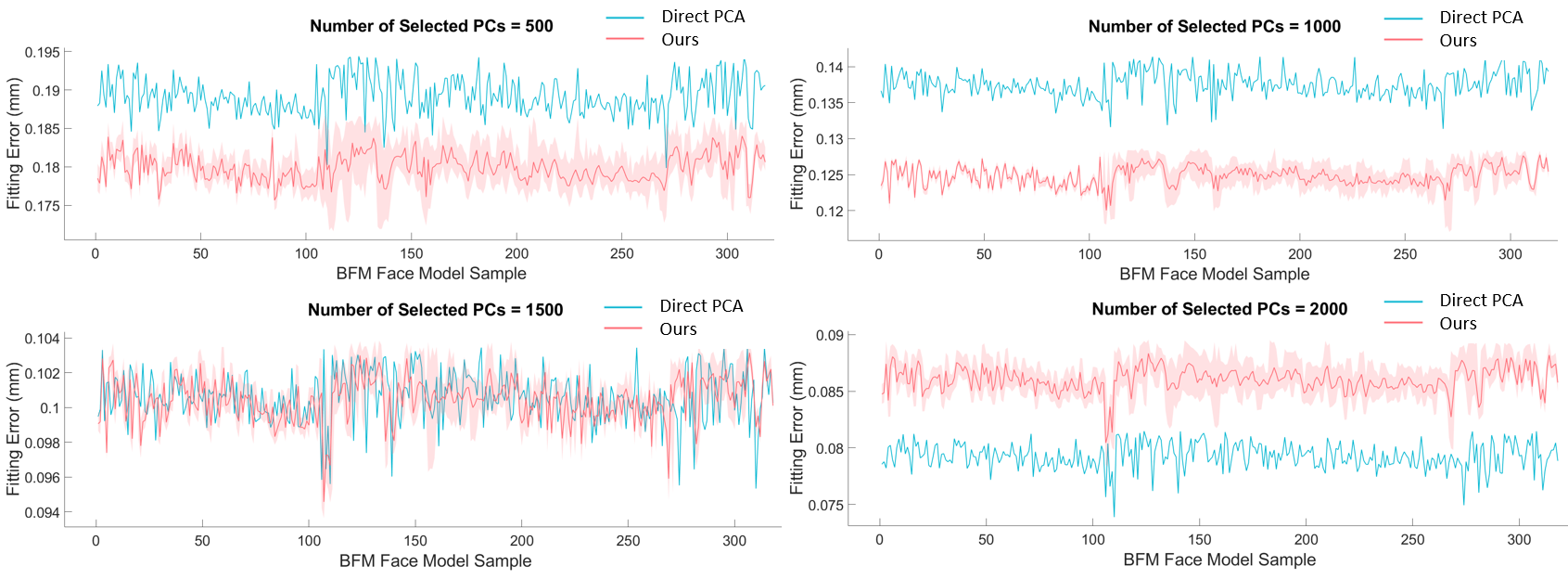}
    \vspace{-3mm}
    \caption{Fitting errors of the BFM face model samples with our stochastic approach and direct PCA. Red shaded regions are the error variations of 10 different runs of our algorithm. Compared with direct PCA, our approach has clear advantages when the used PCs are less than 1500. Note that we use 500 PCs for our facial geometry modeling algorithm in this paper, as we found that the optimization with more than 500 PCs does not yield better results in most of our experiments, possibly due to the optimization difficulty with more parameters and its  sensitivity to noisy RGB-D inputs.}
    \label{fig:comp3dmmdirectpca}
    \vspace{-2mm}
\end{figure*}

In order to validate the effectiveness of our technique, we design a numerical evaluation experiments with the help of BFM 2009 model \cite{paysan20093d}.
Note that the source face models in our dataset are all East Asians, while those in BFM are mostly not Asians.
The domain gap between the two datasets provides us a good benchmark for cross validation (our goal is not to model cross-ethnicity fitting, but use the relative fitting errors between different versions of 3DMM to evaluate their expressive power).
For each BFM basis, we compute two 3D face models using the positive and negative standard deviation values.
A total of 398 BFM face models are obtained in this way.
We register the BFM face models to our mesh topology using Wrap3 software \cite{r3dswrap3}.
We use the extracted PCA basis from our dataset to fit the obtained BFM face models and measure the fitting errors.
Fig. \ref{fig.pcafittingerror} shows the comparison of fitting errors between the augmented basis and the original basis.
If only 100 basis vectors are used, the augmented version has no advantage against the original version.
As the number of basis vectors grows, the augmented version clearly outperforms the original version.
Note that the maximum number of basis vectors of the original version is 200 since there are only 200 source models for 3DMM construction.
For the augmented version, thousands of basis vectors could be obtained since there are over 100,000 models after augmentation.
Since our iterative algorithm emphasizes the expressive power of the principal components with $99.9\%$ cumulative explained variance, most of the expressive capacities are compressed into these components.
In our experiments, we found the final number of the components with $99.9\%$ cumulative explained variance in different runs is generally around 500.
Thus we use 500 basis vectors through this paper.
Fig. \ref{fig.augmentcompare} shows a comparison of the facial geometries obtained using our optimization algorithm in Sec. \ref{sec.optframework} with different versions of PCA basis.

\rev{\subsubsection*{Comparison to Direct PCA \& Stochastic Stability}
We further compare our stochastic approach to direct PCA. Fig. \ref{fig:comp3dmmdirectpca} shows the fitting errors of the BFM face model samples with our method and direct PCA when using different number of principal components (PCs). The results show that when the number of used PCs is less than 1500, our approach consistently outperforms direct PCA. When the number grows up to 1500, direct PCA gradually surpasses our stochastic approach. This can be explained by the intentional design of our stochastic algorithm that emphasizes expressive capacities of the lower dimensions of the basis. Consequently, the expressive power of the components beyond the concerned dimensions are compromised. Note that the red shaded regions in the figure are the error variations of 10 different runs of our stochastic algorithm. The algorithm performs quite stably over different runs.}

\subsubsection*{Relation to Localized 3DMM}

There are some approaches constructing separate 3DMM for each facial region \cite{blanz1999morphable,tena2011interactive,neumann2013sparse}.
The localized 3DMMs obtain more capacities compared with global models by separating the deformation correlations between different facial regions.
The region replacement augmentation in our approach is in the same spirit as localized 3DMM by explicitly generating samples with possible combinations of facial regions from different subjects.
Compared with localized models, our 3DMM avoids online fusion of facial regions and thus is more efficient.
Besides, our perturbation scheme and iterative 3DMM construction algorithm can be applied to localized models to improve their capacities as well.
In this paper, we employ global model for efficiency consideration.

%%%%%%%%%%%%%%%%%%%%%%%%%%%%%%%%%%
%%%%%%%%%%%%%%%%%%%%%%%%%%%%%%%%%%
\section{Facial Reflectance Synthesis}
\label{sec.texturesynth}

\begin{figure*}[t]
    \centering
    \includegraphics[width=0.9\textwidth]{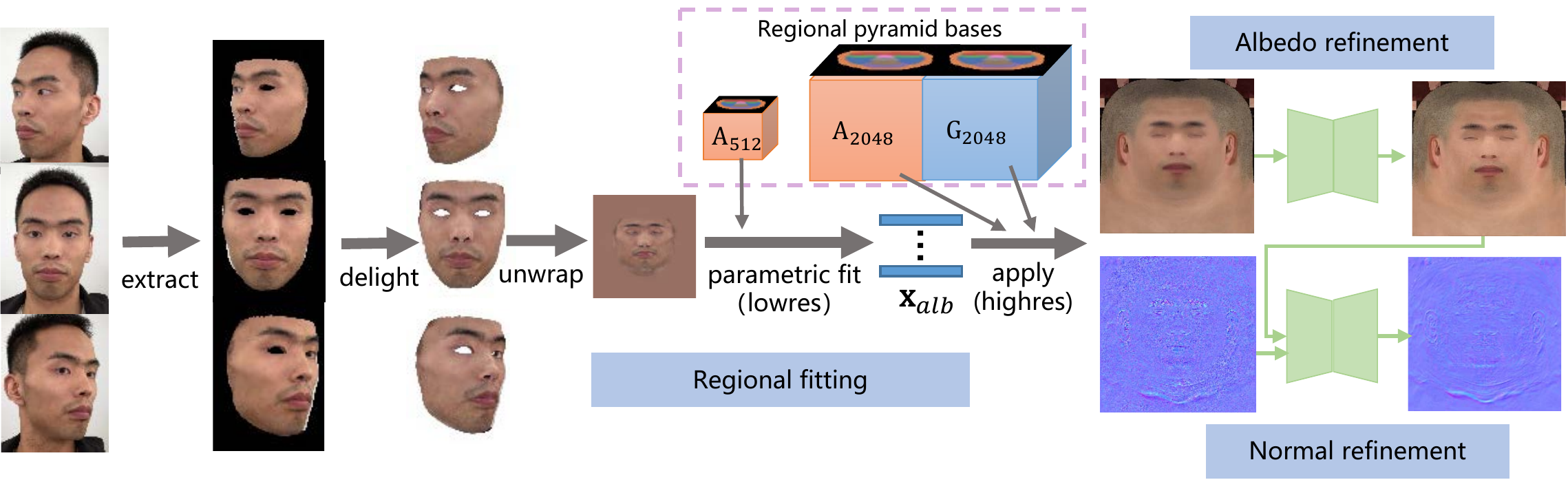}
    \vspace{-3mm}
    \caption{Our albedo/normal map synthesis pipeline. We first extract textures from the source images using the estimated shape and poses from Sec. \ref{sec.optframework}. Then a model-based delighting using the estimated lighting parameters from Sec. \ref{sec.optframework} is applied on the extracted textures, followed by an unwrapping and blending to yield an initial $512 \times 512$ albedo UV-map. Then we use a novel regional fitting approach (Sec. \ref{sec.textregfitting}) to fit the albedo map to get a albedo map and a normal map, both in $2048 \times 2048$ resolution. Finally, the albedo map and normal map are refined with two detail synthesis CNNs, respectively.}
    \label{fig:tex_norm}
    \vspace{-2mm}
\end{figure*}

In this section, we present a our hybrid approach to synthesis high-resolution albedo and normal maps.
We notice that super-resolution based approaches \cite{yamaguchi2018high,lattas2020avatarme} cannot yield high-quality, hair-level details of the eyebrows.
On the other hand, directly synthesizing high-resolution texture maps \cite{saito2017photorealistic} may lead to overwhelming details, which also makes the rendering not realistic.
Our approach addresses the problems with the help of a pyramid-based parametric representation.
Fig. \ref{fig:tex_norm} shows the pipeline of our approach, which we explain as follows.

\subsection{Regional Pyramid Bases}
\label{sec.textureregionbases}

\begin{figure*}[t]
    \centering
    \includegraphics[width=0.9\textwidth]{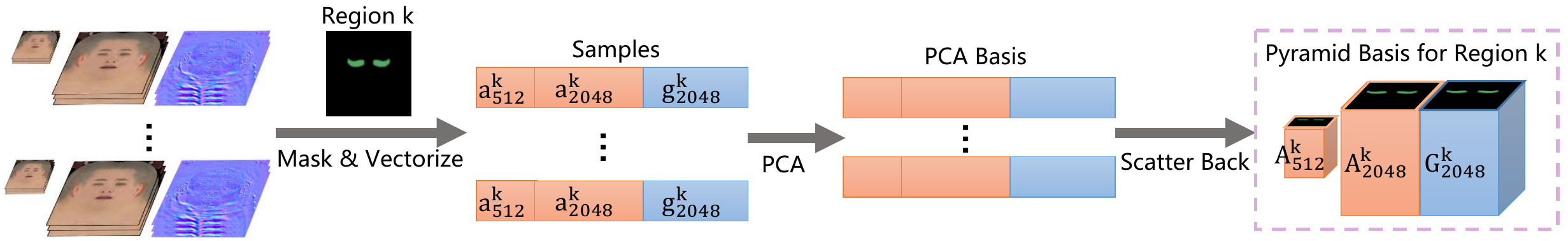}
    \vspace{-3mm}
    \caption{Construction of the regional pyramid bases. \rev{Take the eyebrow region (denoted as region $k$) as an example, the pyramid basis is constructed as follows. First, for each sample in our dataset, we fetch the pixel values in the masked region from the $512$ albedo map, the $2048$ albedo map, and the $2048$ normal map to get $\mathbf{a}_{512}^{k}, \mathbf{a}_{2048}^{k}, \mathbf{g}_{2048}^{k}$, respectively. Then they are vectorized and concatenated together to get a single sample for PCA. Note that each row in the ``samples'' (middle left) is a vectorized sample containing information from all three maps. Thus the PCA process simultaneously captures variations in three maps across resolutions. Finally, the vectorized PCA basis is scattered back into image format by filling the corresponding pixels back into region $k$.}}
    \vspace{-2mm}
    \label{fig:bases}
\end{figure*}

Fig. \ref{fig:bases} illustrates the process to construct our regional pyramid bases.
We first compute image pyramids consisting of two resolutions ($512 \times 512$ and $2048 \times 2048$) for the 200 albedo maps in our
\begin{wrapfigure}{l}{0.18\textwidth}
  \begin{center}
  \vspace{-3mm}
    \includegraphics[width=0.2\textwidth]{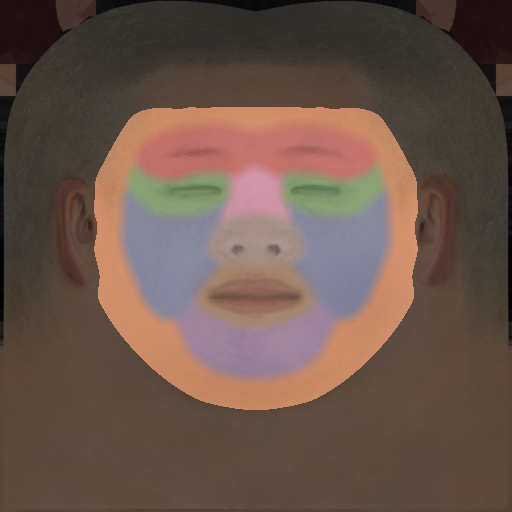}
  \end{center}
  \vspace{-2mm}
\end{wrapfigure}
dataset. We divide facial regions into 8 sub-regions indicated as the different colors in the left UV-map.
The region partitioning is based on the fact that different regions have different types of skin/hair details.
Denote the set of all regions as $\mathcal{K}$.
For each region $k \in \mathcal{K}$, we construct a linear PCA-based blending model.
We define each sample in our dataset as a triplet $(\mathbf{a}_{512}^{k}, \mathbf{a}_{2048}^{k}, \mathbf{g}_{2048}^{k})$, where $\mathbf{a}_{512}^{k}$ stands for $512 \times 512$ albedo map of region $k$, and $\mathbf{a}_{2048}^{k}$ and $\mathbf{g}_{2048}^{k}$ are the albedo map and normal map in $2048 \times 2048$ resolution.
Then the triplet is vectorized into a vector format by fetching and concatenating all pixel values together from the three maps in the region.
Note that during the process, the pixel indices in the three maps are recorded such that the vectorized sample can be ``scatter back\footnote{We use the ``scatter\_nd'' function in Tensorflow as the ``scatter back'' operation.}'' into UV-map format.
For each region $k$, we apply a PCA on the the 200 vectorized samples to get the basis.
Finally, the vectorized basis can be scattered back into UV-map format to obtain the blending basis $\{A_{512}^k, A_{2048}^k, G_{2048}^k\}_{k \in \mathcal{K}}$, where $A_{512}^k \in \mathbb{R}^{n_k \times 199}$ is the low-resolution albedo basis, $A_{2048}^k \in \mathbb{R}^{16n_k \times 199}$ is the high-resolution albedo basis, $G_{2048}^k \in \mathbb{R}^{16n_k \times 199}$ is the high-resolution normal basis. and $n_k$ is the number of pixels within region $k$ in the $512$-resolution.

The constructed regional pyramid bases have several advantages compared with conventional bases.
First, the expressive capacity is larger than global linear bases, while each region can be processed individually to accelerate the runtime.
Second, the bases capture variations in both albedo and normal map.
Third, the incorporated multiple resolutions can emphasize more structural information in the extracted bases.
With the pyramid basis, we can perform parametric fitting on the low resolution, and directly apply the same parameters on high-resolution bases to obtain high-resolution albedo and normal maps.
This not only reduces computation, but also generally yields higher-quality results than directly fitting on high resolution.
Note that the albedo basis employed in our geometric fitting procedure (Sec. \ref{sec.optframework}) is the conventional global model in $512$-resolution, while the bases used in this section dedicated for reflectance synthesis are the regional pyramid model.
The reason is that we found the less powerful albedo basis would make the optimization constraints more imposed on the geometric parameter estimation rather than texture parameter estimation.
Otherwise more powerful albedo basis tends to result in overfitted textures but underfitted geometries.

\subsection{Regional Fitting}\label{sec.textregfitting}
%\textbf{Fitting process}.

Since the albedo parameters $\mathbf{x}_{alb}$ obtained in Sec. \ref{sec.optframework} are based on conventional global bases, the resulting albedo maps are not satisfactory due to limited expressive power.
Here we directly extract textures from the source images using the estimated shape and poses from Sec. \ref{sec.optframework}.
Then a model-based delighting using the estimated lighting parameters from Sec. \ref{sec.optframework} is applied on the extracted textures, followed by an unwrapping and blending to yield an initial $512 \times 512$ albedo map $I_{init}$.
We use the $512$-resolution regional bases $\{A_{512}^k\}_{k \in \mathcal{K}}$ to fit the initial albedo map:
\begin{equation*}
    L(\mathbf{x}_{alb}) = \| I_{fit}(\mathbf{x}_{alb}) - I_{init} \|_{2} + \omega_{tv} f_{tv}(I_{fit}(\mathbf{x}_{alb})) + \omega_{alb} \| \mathbf{x}_{alb} \|_{2}^{2},
\end{equation*}
where $I_{fit}(\mathbf{x}_{alb}) = \sum_{k \in \mathcal{K}} A_{512}^k \mathbf{x}_{alb}^{k}$, $f_{tv}$ denotes the total variation function, $\omega_{tv} = 0.0001$ and $\omega_{alb} = 0.001$.
Note that the total variation term is essential to eliminate the artifacts in the resulting albedo maps near boundaries between regions.
After obtaining $\mathbf{x}_{alb}$, we can directly compute a high-resolution albedo map $\mathbf{\hat{a}}_{2048}$ and a normal map $\mathbf{\hat{g}}_{2048}$ as
\begin{align}
    \label{eq:computehighresfit}
    \mathbf{\hat{a}}_{2048} &= \sum_{k \in \mathcal{K}} A_{2048}^k \mathbf{x}_{alb}^{k} \nonumber \\
    \mathbf{\hat{g}}_{2048} &= \sum_{k \in \mathcal{K}} G_{2048}^k \mathbf{x}_{alb}^{k} \nonumber.
\end{align}

With the help of regional pyramid bases, different types of skin/hair details in different regions can be separately preserved via the high-resolution bases, while the fitting process on low resolution makes the algorithm focus on major facial structures, e.g., the shape of the eyebrows and lips.
Since the parametric representation is based on linear blending model, the results are usually over-smoothed (see Fig. \ref{fig:textureablationcloseup}).
We next present our detail synthesis step to refine the albedo and normal maps.

\subsection{Detail Synthesis}
%\textbf{Detail-enhancing network}.

We adopt two refinement networks to synthesize details for albedo and normal map respectively.
The refinement networks employ the architecture of a GAN-based image translation model, \emph{pix2pix} \cite{isola2017image}.
As shown in Fig. \ref{fig:tex_norm}, for albedo refinement, the network takes the fitted $2048$-resolution albedo map as input and outputs a refined albedo map in the same resolution.
For normal refinement, the refined albedo map and the fitted normal map are concatenated along channel dimension. The refinement network takes the concatenation as inputs and outputs a refined normal map.

During training, we first use facial region replacement and skin color transfer \cite{reinhard2001color} to augment the 200 high-quality albedo/normal maps (from the dataset for constructing the 3DMM) into 4000 maps, which serve as ground-truth supervision for training the two networks.
Then we perform regional fitting (Sec. \ref{sec.textregfitting}) on the 4000 maps to get the fitted albedo/normal maps, which serve as inputs of the networks during training.
We only use the facial regions out of the whole UV maps for computing training losses.
Similar to \emph{pix2pix} \cite{isola2017image}, we keep $L1$ loss and GAN loss in both networks.
For albedo refinement, we additionally apply total variation loss to reduce artifacts and improve skin smoothness.
The weights for $L1$, GAN and total variation losses are  $100$, $1$, $0.001$.
For normal refinement, we additionally employ a pixel-wise cosine distances between the predictions and ground-truth maps to increase the accuracy of normal directions.
The weights for $L1$, GAN and cosine distance losses are $100$, $1$, $0.001$.
The networks are trained with Adam optimizer for $75000$ iterations.

\subsubsection*{Relation to Existing Approaches}

There are three recent CNN-based approaches that can be adopted to synthesize high-resolution facial UV-maps, which are Yamaguchi et al. \shortcite{yamaguchi2018high}, GANFIT \cite{gecer2019ganfit}, and AvatarMe \cite{lattas2020avatarme}.
However, these approaches cannot produce satisfactory results in our case.
GANFIT \cite{gecer2019ganfit} needs about $50$ times more training data than ours to train a GAN as the nonlinear parametric representation of texture maps.
In their work, the $10,000$ texture maps are obtained using unwrapped photos, where shadings and specular highlights are not removed.
In our system, the $200$ albedo maps and normal maps are created with very high-quality artistic efforts, where shadings and specular highlights are completely removed and hair-level details are preserved.
It is rather difficult to extend our data amount to theirs while keeping such high data quality.
Regarding the other two approaches, we also tried super-resolution based network as in Yamaguchi et al. \shortcite{yamaguchi2018high} and pure CNN-based synthesis in AvatarMe \cite{lattas2020avatarme}, and found the results obtained with their approaches are generally inferior to ours.
We present some comparison in Sec. \ref{sec.expresulttexture}.

%%%%%%%%%%%%%%%%%%%%%%%%%%%%%%%%%%
%%%%%%%%%%%%%%%%%%%%%%%%%%%%%%%%%%
\section{Full Head Rig Creation and Rendering}\label{sec.fullheadrig}
% head completion
% hair classification
% expression rigging
\subsubsection*{Head Completion}
Although the scanned 200 models in out dataset are full head models, there are usually no reliable geometric constraints beyond facial regions for a RGB-D selfie user.
We employ an algorithm to automatically complete a full head model given the recovered facial model.
The regions involved in our algorithm are:
\vspace{-1mm}
\begin{itemize}
  \item A: facial region;
  \item B: back head region;
  \item C: intermediate region;
  \item D: overlapped region between A and C.
\end{itemize}
\vspace{-1mm}
\begin{figure}[H]
\centering
\vspace{-2mm}
\includegraphics[width=0.46\textwidth]{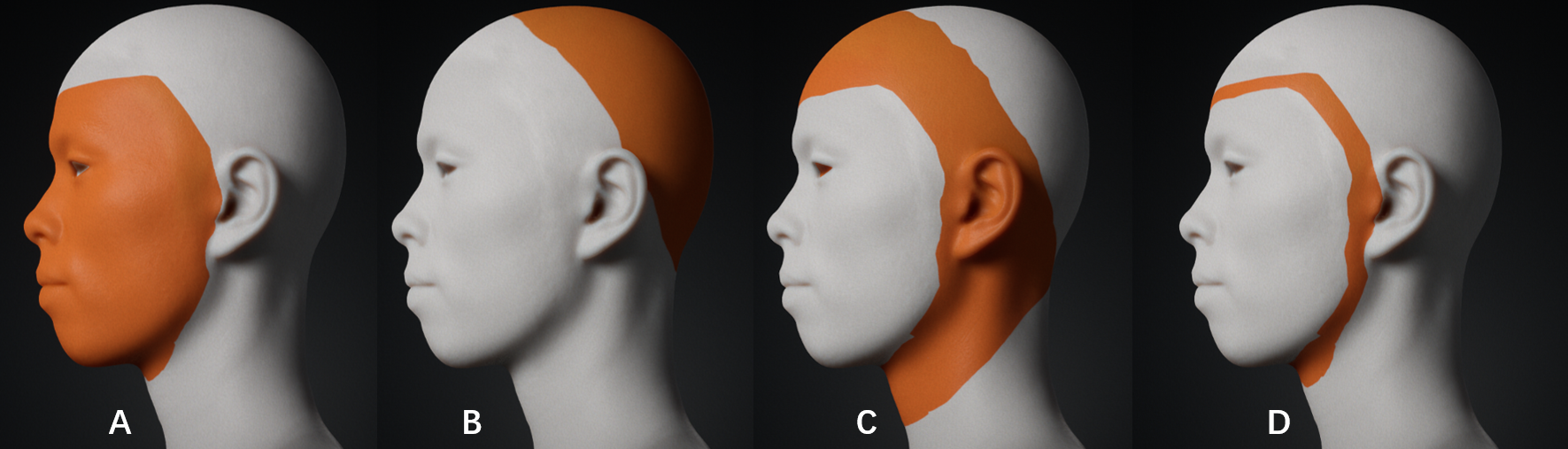}
\vspace{-2mm}
\label{Fig.11}
\end{figure}
Our goal is to compute a full head shape such that region A matches the facial shape and region B matches a reference back head shape.
The reason of using a reference shape for back head region B is it can further ease the difficulties to attach accessories like hair models.
To this end, we construct a head morphable model using only regions B $\cup$ C of the 200 source models.
Note that this model does not need strong expressive power as face models, thus we do not employ the technique in Sec. \ref{sec.3dmmaug} but directly use PCA to extract basis.
Given the recovered facial shape of a user, we apply a ridge regression similar to Sec. \ref{sec.initfitting} to get the head 3DMM parameters, using the constraints of region B $\cup$ D.
Then the full head model is obtained by combining the resulting shape (B $\cup$ C) with facial region A.

\begin{figure}[h]
\centering
\vspace{-3mm}
\includegraphics[width=0.45\textwidth]{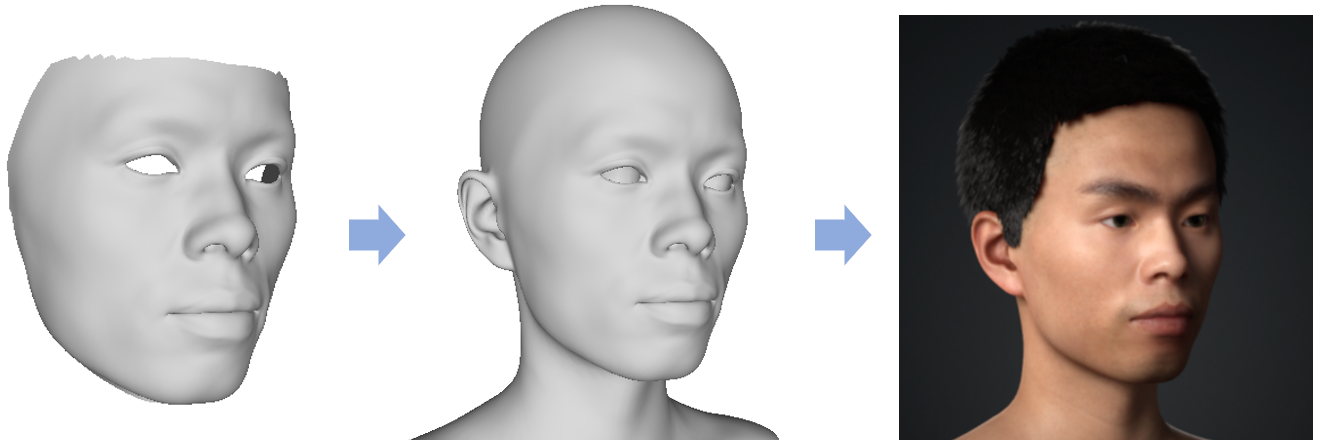}
\vspace{-5mm}
\label{Fig.add_accessories}
\end{figure}

\subsubsection*{Accessories}

We perform a hairstyle classification on the user's front photo (using a MobileNet \cite{howard2017mobilenets} image classification model trained on labeled front photos) and attach the corresponding hair model (created by artists in advance) to the head model according to the predicted hairstyle label.
There are in total 30 hairs models in different hairstyles in our system (see supplementary materials).
For eyeballs, we use template models and calculate the positions and scales based on reference points on the head model.
For teeth, we employ an upper teeth model and a lower teeth model.
The upper teeth model is placed according to reference points near nose, and it remains still when facial expression changes.
The lower teeth model is placed according to reference points on the chin.
When mouth opens or closes, the lower teeth model moves with the chin.
Note that the accessory models are not the focus of this work, their modeling and animation can be found in dedicated research work \cite{zoss2018empirical,wu2016model,berard2019practical,berard2016lightweight,zoss2019accurate,velinov2018appearance}.

\subsubsection*{Expression Rigging}
We adopt a simple approach similar to Hu et al. \shortcite{hu2017avatar} to transfer generic FACS-based blendshapes to the target model to obtain expression blendshapes.
Note that our approach can be extended to further acquire user's expression data and construct personalized blendshapes like Ichim et al. \shortcite{ichim2015dynamic}.

\subsubsection*{Rendering}
The recovered full head mesh model, as well as the high-quality albedo map and normal map, can be rendered with any physically based renderer.
In this work, we show rendered results using Unreal Engine 4 (UE4), \rev{with a skin PBR material template simplified from the official skin PBR material template provided by the engine \shortcite{ue4digitalhuman}.
The skin PBR material details can be found in the supplementary materials.}

%%%%%%%%%%%%%%%%%%%%%%%%%%%%%%%%%%
%%%%%%%%%%%%%%%%%%%%%%%%%%%%%%%%%%
\section{Results and Evaluation}\label{sec.expresult}

\begin{table}[t]
    \centering
    \begin{tabular}{c|c}
    \hline
        \textbf{Processing Step} & \textbf{Runtime} \\
        \hline
        \hline
        Landmark Detection & -- \\
        Coarse Screening & -- \\
        Bilateral Filtering & -- \\
        Frame Selection & 0.2s \\
        Initial Model Fitting & 0.1s \\
        Initial Texture & 0.5s \\
        Optimization & 10s \\
        Regional Parametric Fitting & 1.5s \\
        Detail Synthesis & 1s \\
        Head Completion & 0.5s \\
        Accessories & 0.1s \\
        Expression Rigging & 1s\\
        \hline
        Total & $\sim$15s \\
        \hline
    \end{tabular}
    \vspace{1mm}
    \caption{The runtime for each step in our system. Note that the first three steps are computed during data acquisition and thus do not need additional processing time. GPU and multi-thread CPU are used.}
    \vspace{-4mm}
    \label{tab:stepruntime}
\end{table}

\begin{table}[t]
    \centering
    \begin{tabular}{c|c|c|c}
    \hline
        \textbf{Avatar Creation} & \textbf{Acquisition} & \textbf{Processing} & \textbf{Manual} \\
        \textbf{System} & \textbf{Time} & \textbf{Time} & \textbf{Interaction} \\
        \hline
        \hline
        \cite{ichim2015dynamic} & 10 minutes & $\sim$1 hour & 15 minutes \\
        \cite{cao2016real} & 10 minutes & $\sim$1 hour & needed  \\
        \cite{hu2017avatar} & $<$1 second & $\sim$6 minutes & --  \\
        Ours & $<$10 seconds & $\sim$15 seconds & -- \\
        \hline
    \end{tabular}
    \vspace{1mm}
    \caption{Time comparison with other avatar creation systems.}
    \vspace{-5mm}
    \label{tab:timecompare}
\end{table}

\subsection{Acquisition and Processing Time}

The selfie data acquisition typically takes less than 10 seconds (200-300 frames).
The total processing time after data acquisition is about 15 seconds.
Note that some of the processing steps like real-time landmark detection, coarse screening, and bilateral filtering can be computed on the smartphone client while the user is taking selfie.
The preprocessed data are streamed to a server via WiFi during acquisition.
The rest steps of the processing are computed on the server.
Table \ref{tab:stepruntime} shows the runtime on our server with an Nvidia Tesla P40 GPU and an Intel Xeon E5-2699 CPU (22 cores).
Note that the frame selection and expression blendshape generation are implemented with multi-thread acceleration and the total runtime is largely reduced thanks to parallel processing.
Table \ref{tab:timecompare} shows a time comparison with other avatar creation systems.
In terms of total acquisition and processing time, our system provides a convenient and efficient solution for users to create high-quality digital humans.

\begin{figure}[t]
    \centering
    \includegraphics[width=0.47\textwidth]{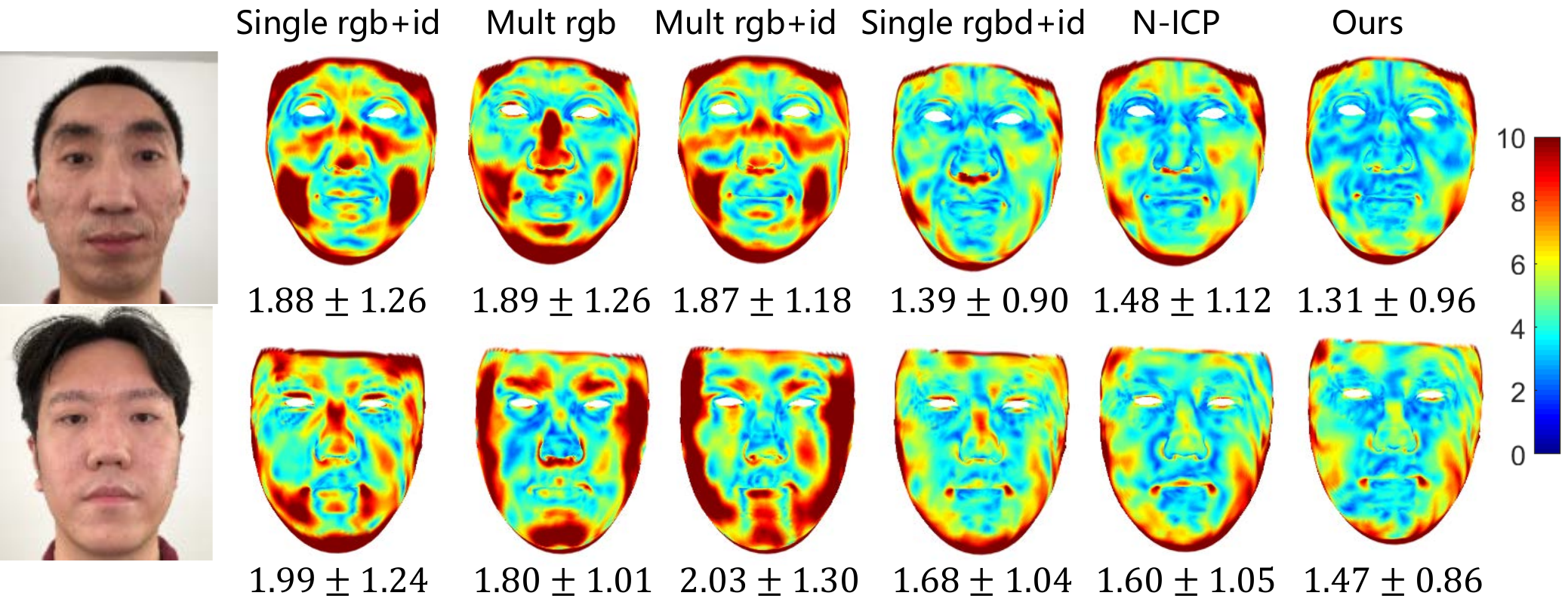}
    \vspace{-3mm}
    \caption{Error maps (in mm) for different variants of our approach.}
    \label{fig:error-map}
\end{figure}

\begin{figure}[t]
    \centering
    \vspace{-1mm}
    \includegraphics[width=0.47\textwidth]{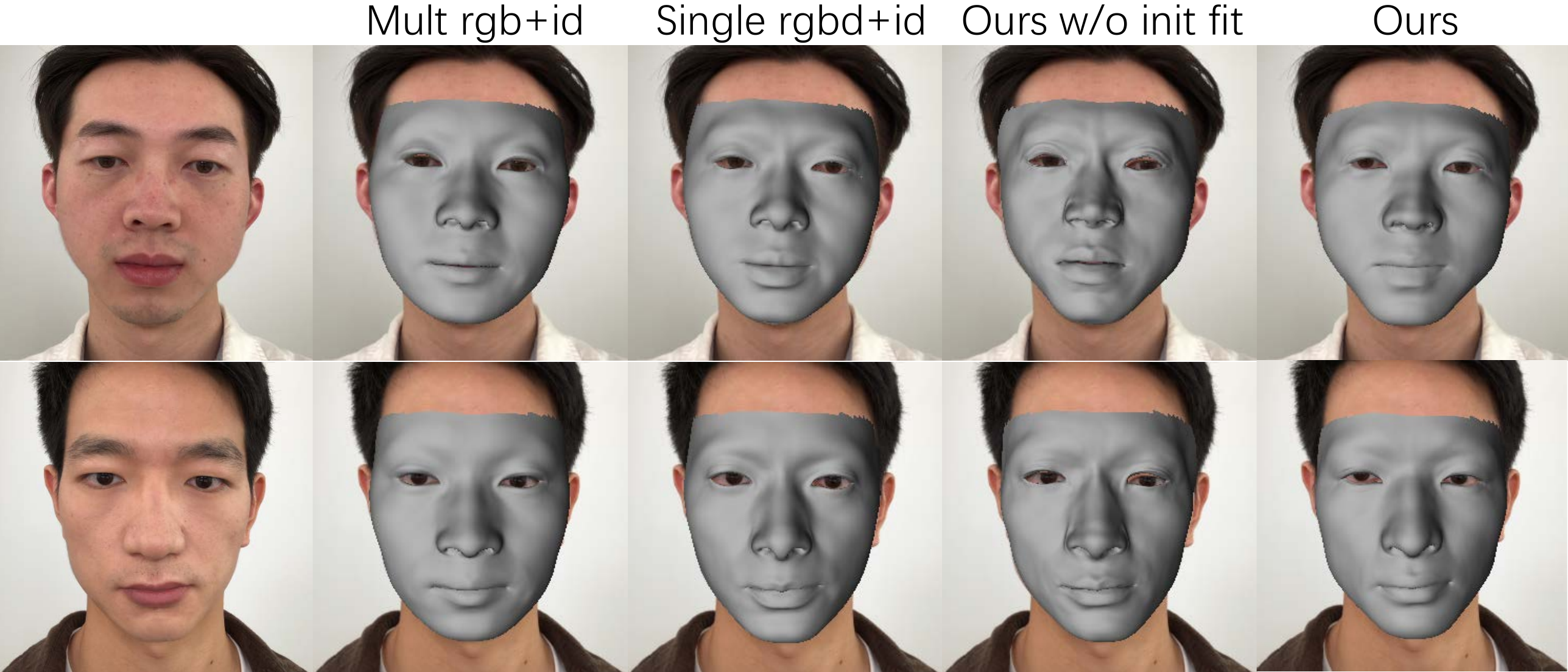}
    \vspace{-3mm}
    \caption{Visual comparison for different variants of our approach. The results obtained with multiview RGB-D data and identity loss (ours) are generally more faithful than other results. We also show the results obtained without initial model fitting, which are usually flawed.}
    \label{fig:ablation}
\end{figure}

\begin{figure*}[t]
    \centering
    \includegraphics[width=0.995\textwidth]{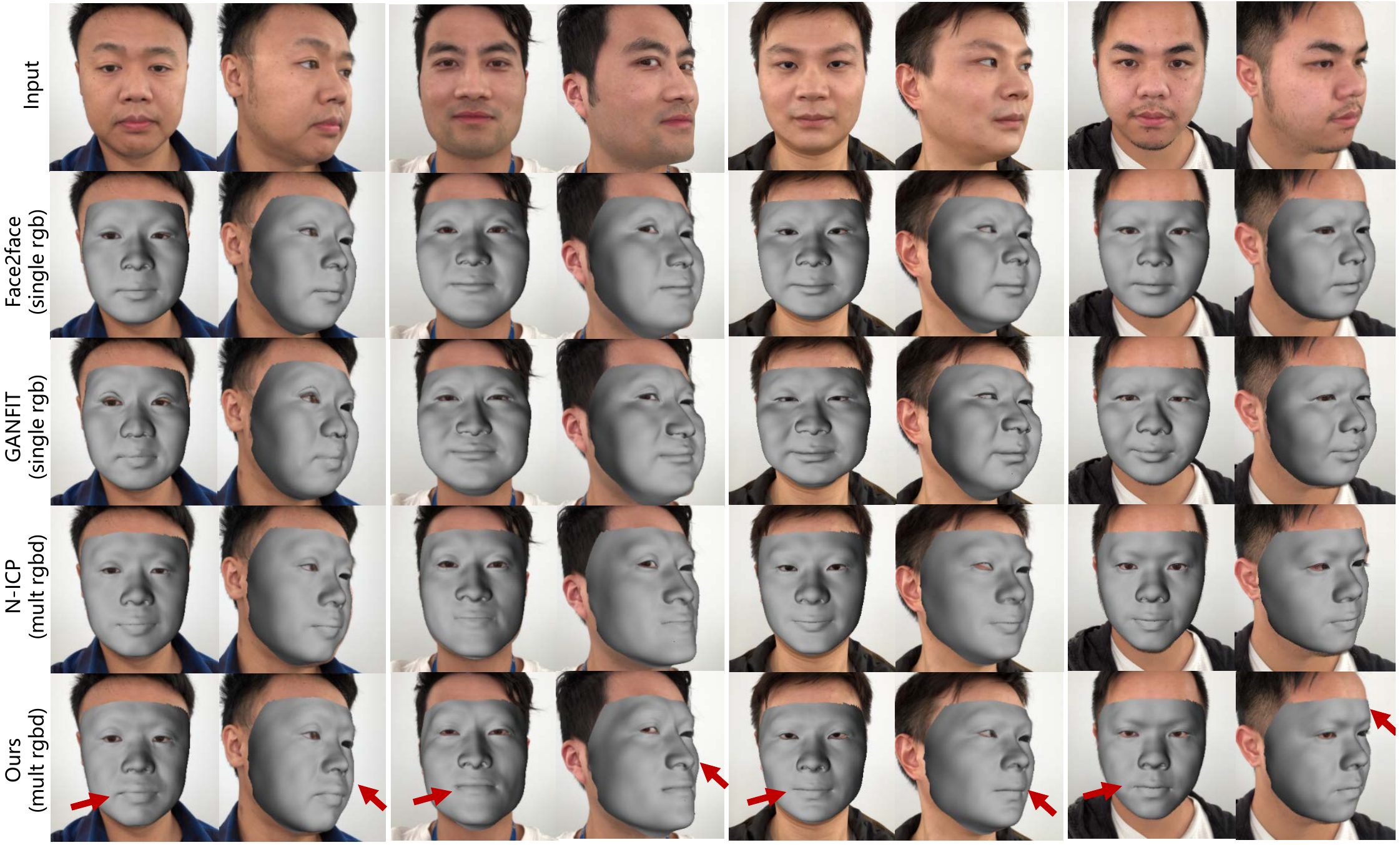}
    \vspace{-3mm}
    \caption{\revv{Visual comparison with state-of-the-art approaches}. As pointed out by the red arrows, our method is able to generate face geometries with more accurate cheek silhouettes and more faithful mouth shapes to the input photos. In comparison, the mouth shapes obtained by N-ICP lack personalized features and are similar to each other among all the subjects. For fair comparison, all the results are obtained with our 3DMM.}
    \label{fig:shape}
    \vspace{-1mm}
\end{figure*}

\begin{figure*}[t]
    \centering
    \includegraphics[width=0.995\textwidth]{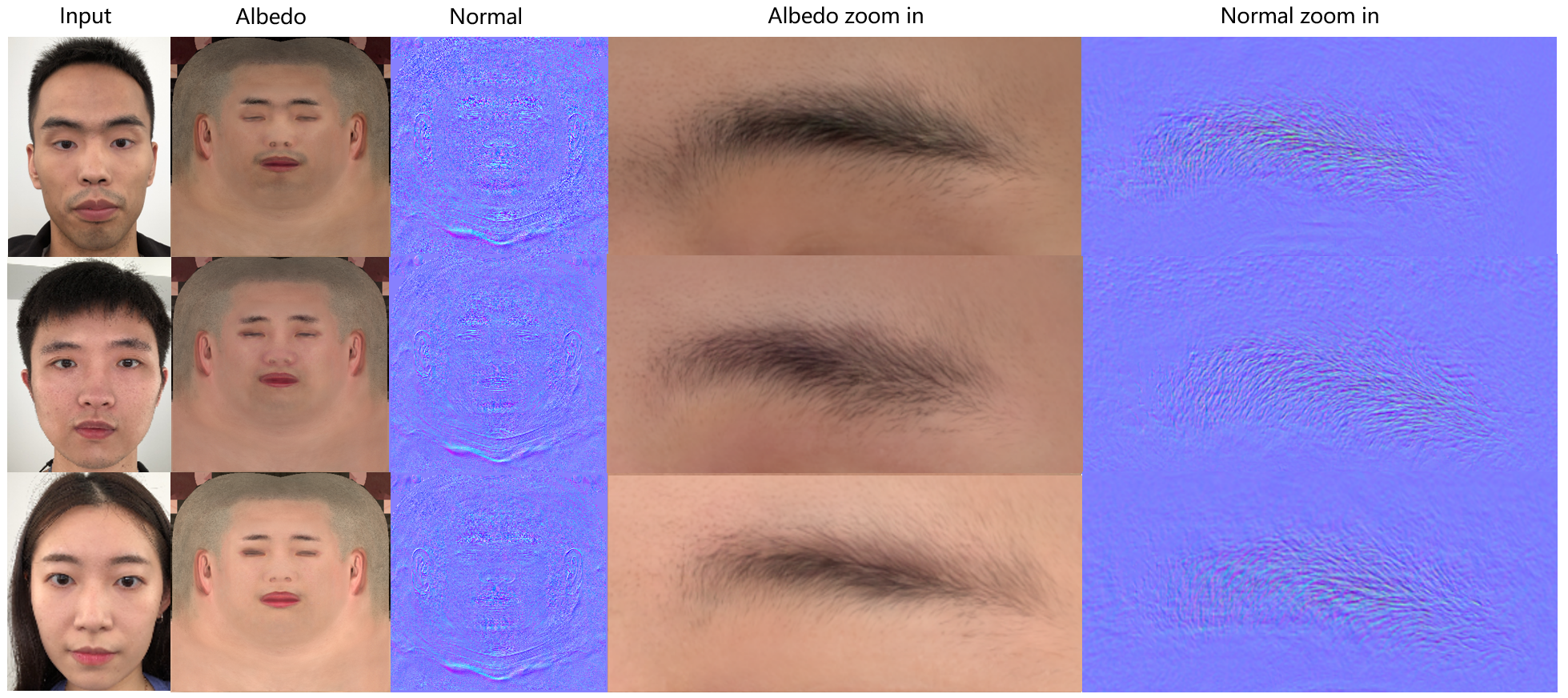}
    \vspace{-3mm}
    \caption{Examples of our synthesized albedo and normal maps.}
    \label{fig:detail-zoom}
    \vspace{-1mm}
\end{figure*}

\begin{figure*}[t]
    \centering
    \vspace{-1mm}
    \includegraphics[width=0.95\textwidth]{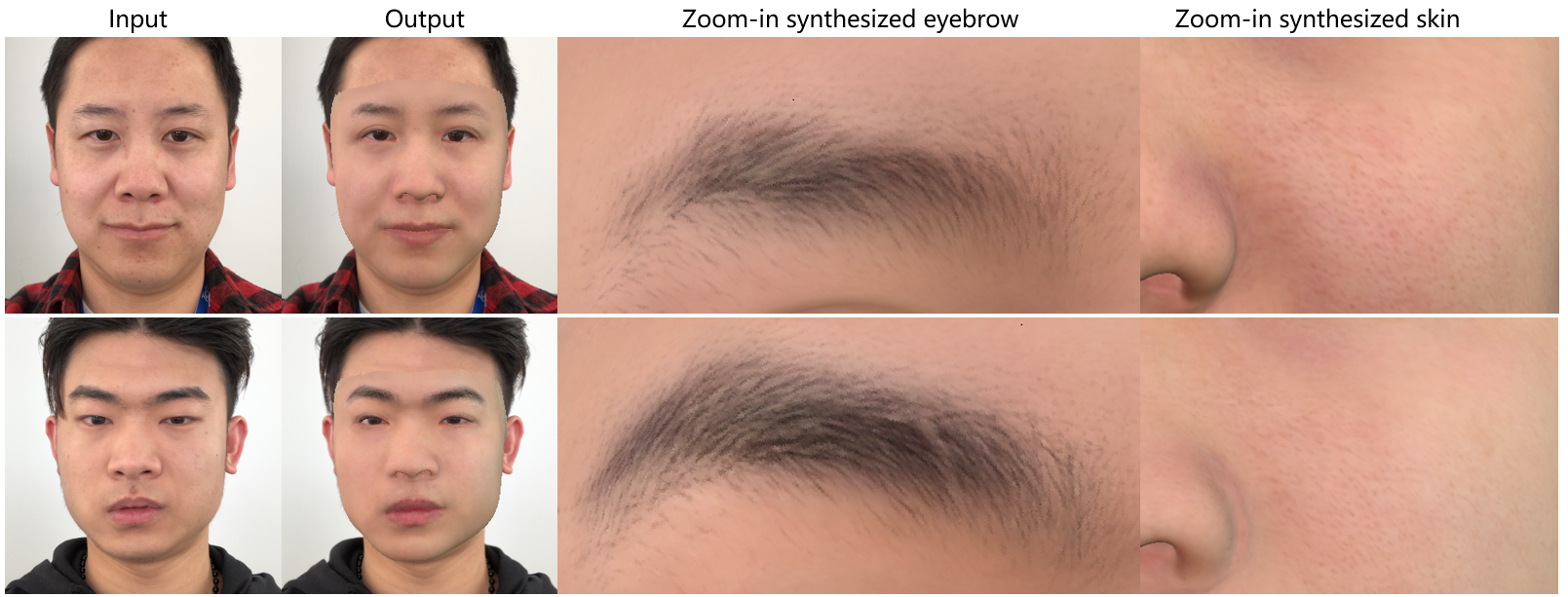}
    \vspace{-3mm}
    \caption{Overlay results with our synthesized albedo maps. Note that the eyebrow and mouth shapes in our albedo maps are faithful to input photos.}
    \vspace{-1mm}
    \label{fig:detail-in-image-space}
\end{figure*}

\begin{figure*}[t]
    \centering
    \vspace{-1mm}
    \includegraphics[width=0.95\textwidth]{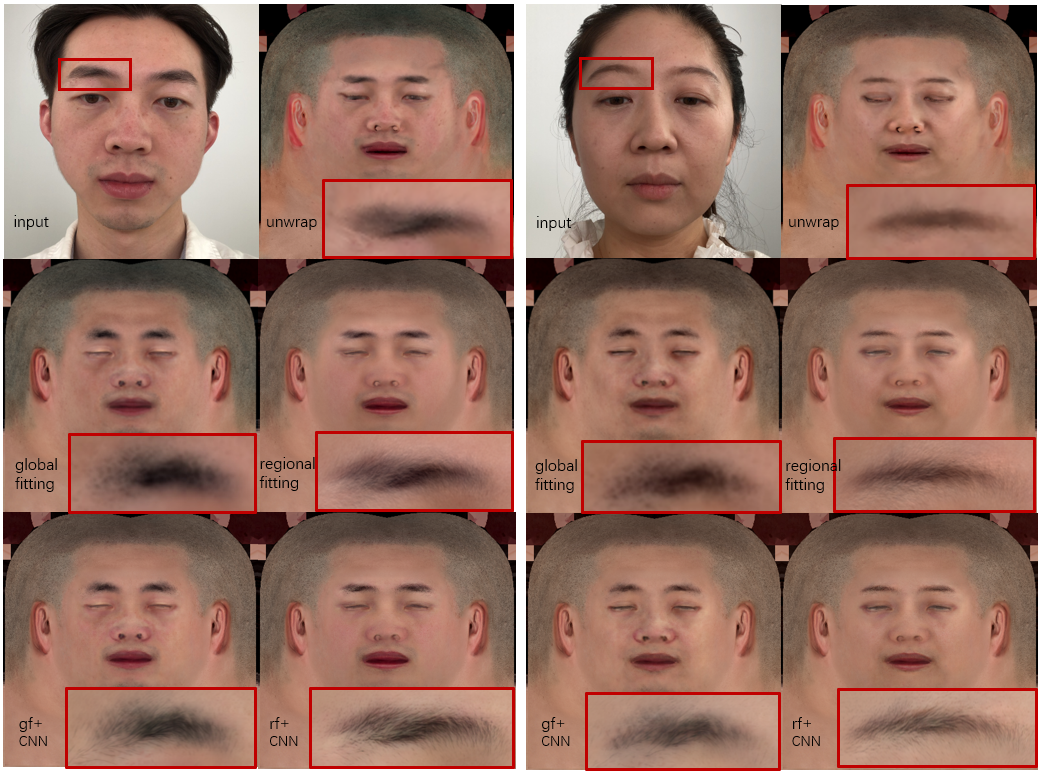}
    \vspace{-3mm}
    \caption{\rev{Ablation results of intermediate or alternative steps of our albedo synthesis pipeline. The global fitting results are obtained from our 3DMM optimization in Sec. \ref{sec.optframework}. The recovered facial structures (like the eyebrow shapes) are not faithful to inputs due to limited capacities of the global texture basis. In contrast, our regional fitting results are much more faithful. The final refinement CNN further improves the realism by adding more hair/pore details.}}
    \vspace{-1mm}
    \label{fig:textureablation}
\end{figure*}

\begin{figure*}[t]
    \centering
    \includegraphics[width=0.995\textwidth]{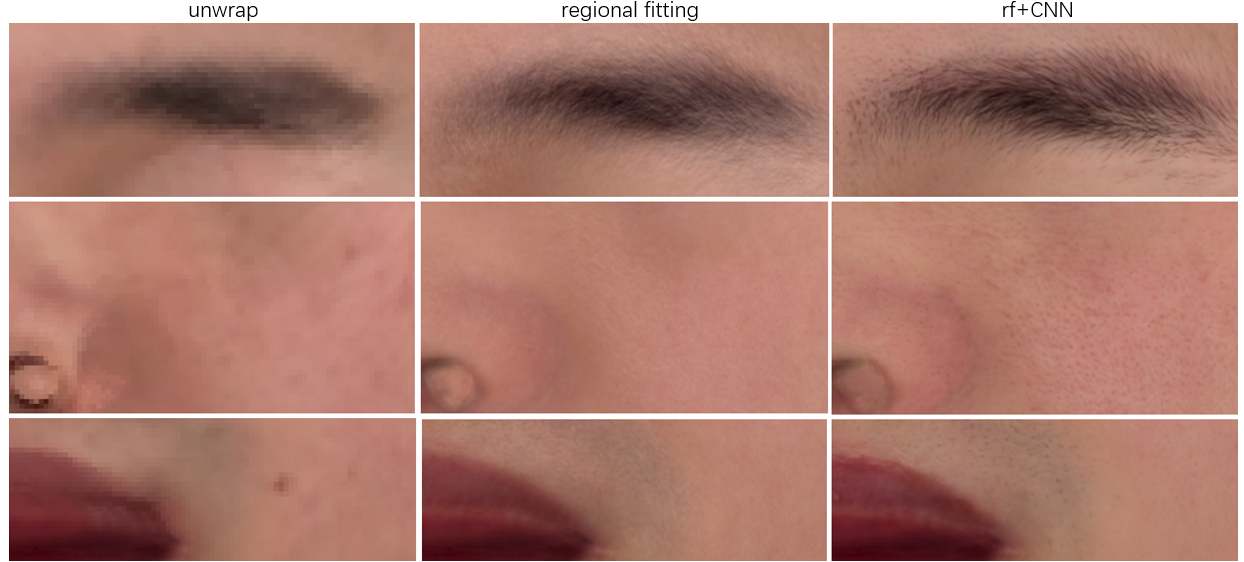}
    \vspace{-3mm}
    \caption{\rev{Close-ups of the synthesized albedo results for the left subject in Fig. \ref{fig:textureablation}.}}
    \label{fig:textureablationcloseup}
\end{figure*}

\begin{figure*}[t]
    \centering
    \includegraphics[width=0.995\textwidth]{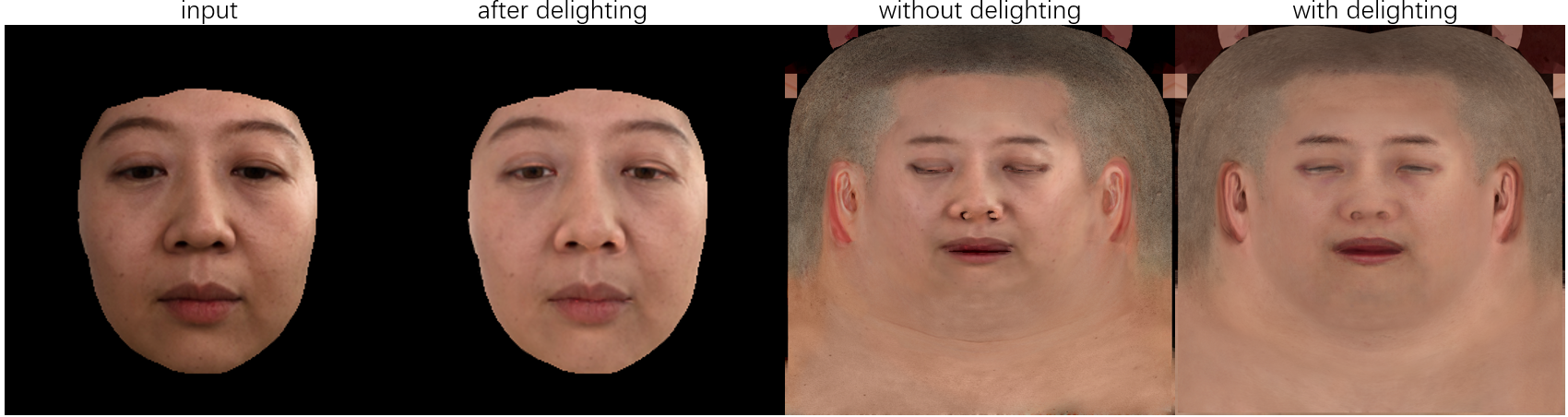}
    \vspace{-3mm}
    \caption{\revv{Ablation results of our albedo synthesis pipeline with/without the delighting step.}}
    \label{fig:delightablation}
\end{figure*}

\subsection{Quality of Facial Geometry}
\label{sec.expresultgeometry}

We evaluate the quality of our recovered facial shapes in extensive experimental settings.
The experiments include quantitative and qualitative comparisons of different variants of our approach and existing methods such as Face2Face \cite{thies2016face2face,hu2017avatar,yamaguchi2018high}, GANFIT \cite{gecer2019ganfit,lattas2020avatarme}, N-ICP \cite{weise2011realtime,bouaziz2016modern}, etc.
Note that all the results are obtained with our 3DMM basis for fair comparisons.

\subsubsection*{Quantitative Evaluation}
We use the same workflow as the production of our dataset to manually create the ground-truth models of two users.
Since the ground-truth obtained in this way is very expensive, we only perform numerical evaluations on the two models to get quantitative observations.
\revv{The corresponding geometries recovered from RGB-D selfie data with different variants of our approach are evaluated.}
The results are in Fig. \ref{fig:error-map}.
It can be seen from the results that our method yields the lowest mean errors, closely followed by N-ICP and the single-view variants of our method (\emph{single rgbd+id}).
Compared with N-ICP, our method performs better on detailed facial geometries near eyes, nose, and mouth.
It conforms to our motivation that appearance constraints (photo loss and identity perceptual loss) help capture more accurate facial features.

\subsubsection*{Qualitative Evaluation}

Fig. \ref{fig:ablation} shows two examples of the shape models obtained with different variants of our method.
The version with multiview RGB-D data generally outperforms other variants.
Besides, as shown in the figure, random initialization of our optimization can lead to flawed models.
The initial fitting in Sec. \ref{sec.initfitting} improves the system robustness.
We further show results comparisons between our method with Face2Face \cite{thies2016face2face,hu2017avatar,yamaguchi2018high}, GANFIT \cite{gecer2019ganfit,lattas2020avatarme}, and N-ICP \cite{weise2011realtime,bouaziz2016modern} in Fig. \ref{fig:shape}.
In general, both our method and N-ICP can reconstruct more accurate facial shapes than the methods using only RGB data (Face2face and GANFIT). This can be clearly observed from the silhouettes near the cheek region in the results.
Watching more closely, our method can recover more faithful and personalized facial shapes than N-ICP, especially near mouth region.
The mouth shapes obtained with N-ICP are similar among all the faces, while our results preserve personalized mouth features and are more faithful to the photos.
This can be explained by the additionally incorporated photometric loss and identity loss in our method.

\subsection{Quality of Facial Reflectance}
\label{sec.expresulttexture}

\subsubsection*{Results}

Our method can produce albedo and normal maps with high-quality, realistic details while preserving major facial features of the users.
Fig. \ref{fig:detail-zoom} shows several examples of our obtained albedo and normal maps.
Hair-level details in the albedo and normal maps are clearly visible.
Fig. \ref{fig:detail-in-image-space} shows the overlay results with synthesized albedo maps.
The major facial features are consistent between the synthesized albedo maps and the input photos, especially in the eyebrow region.
More importantly, the hair-level details near eyebrow region and the pore-level skin details are clearly visible.
Note that the input RGB-D images in our method are in $640 \times 480$ resolution, where actual skin micro/meso-structures and hair-level details are hardly visible (see Fig. \ref{fig:detail-in-image-space} left column).
The synthesized skin/hair details by our approach are actually plausible hallucination, which is critical for realistic rendering of digital humans.

\rev{\subsubsection*{Ablation Study}
Since our albedo synthesis pipeline consists of several major steps, we show the ablation results of intermediate or alternative steps of our pipeline in Fig. \ref{fig:textureablation}.
The unwrapped textures are typically blurry and dirty due to low-resolution inputs and imperfect lighting removal.
The global fitting results cannot faithfully preserve major facial structures (like the eyebrow shapes) due to limited expressive power of global texture basis.
Our regional fitting can recover more accurate facial structures, while lacking hair/pore details.
The final refinement CNN supplements hallucinated high-quality, realistic details to the albedo maps.
Fig. \ref{fig:textureablationcloseup} shows several additional close-ups of the intermediate results.
\revv{Fig. \ref{fig:delightablation} shows an example of the results obtained with/without delighting.}
}

\subsubsection*{Pix2pix vs pix2pix-HD}
\begin{figure}[t]
    \centering
    \includegraphics[width=0.48\textwidth]{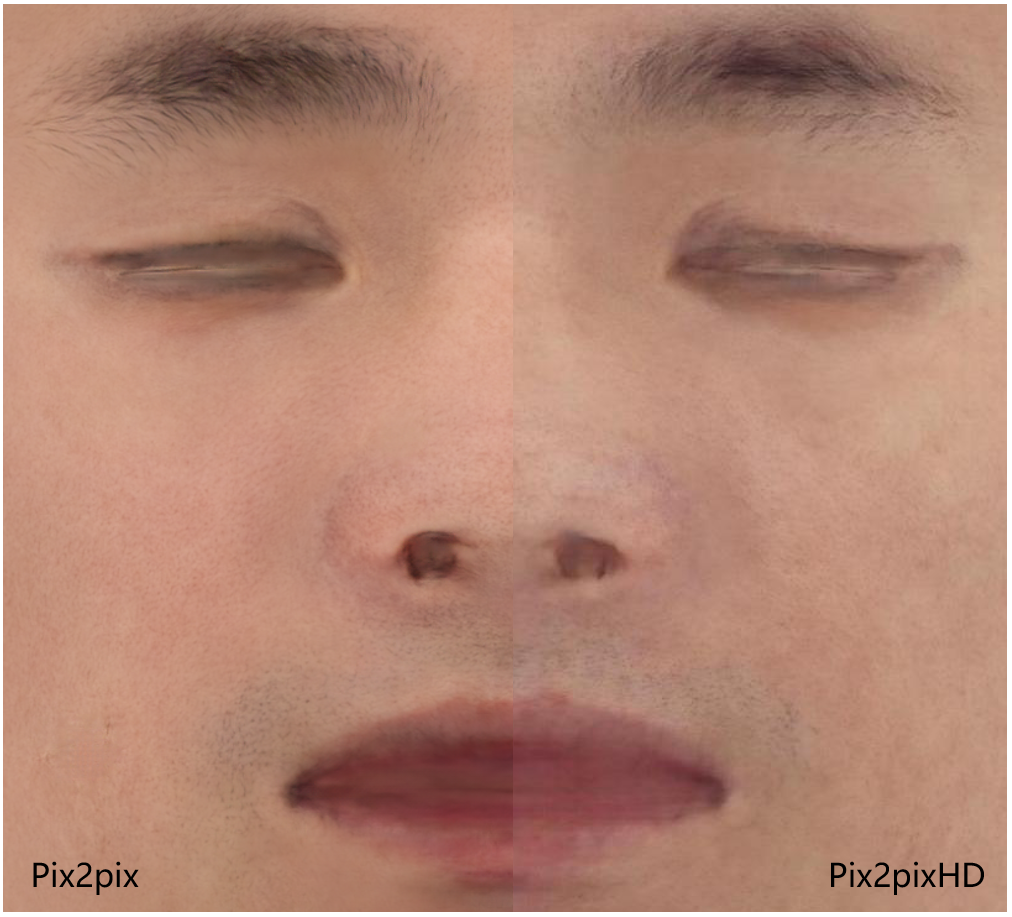}
    \vspace{-6mm}
    \caption{Comparison between pix2pix and pix2pix-HD. The results obtained with pix2pix-HD are generally with uneven skin colors and obvious artifacts.}
    \vspace{-1mm}
    \label{fig:pix2pix-HD}
\end{figure}

We experimented with two variants of our detail synthesis CNNs, i.e., \emph{pix2pix} \cite{isola2017image} and \emph{pix2pix-HD} \cite{wang2018pix2pixHD}.
We found \emph{pix2pix-HD} is more difficult to train and the results are generally worse than \emph{pix2pix} (see Fig. \ref{fig:pix2pix-HD} for an example), possibly due to the small amounts of training data.

\begin{figure*}[t]
    \centering
    \vspace{-1mm}
    \includegraphics[width=0.95\textwidth]{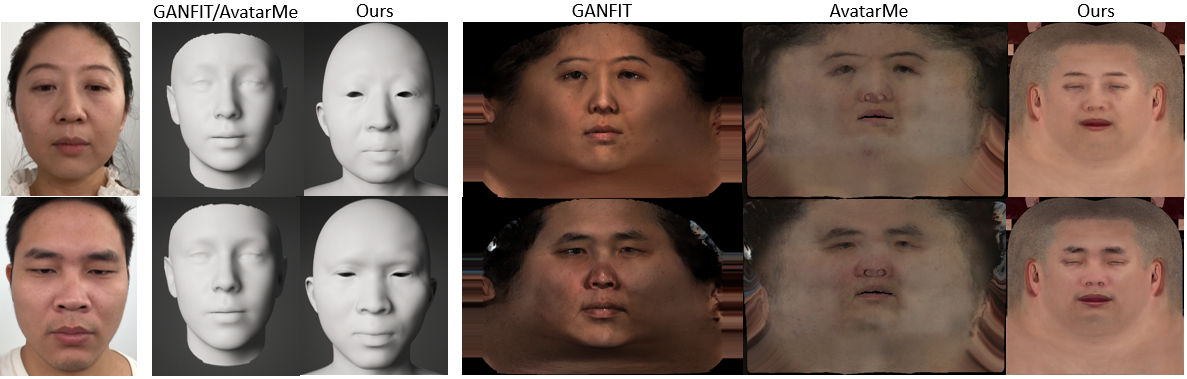}
    \vspace{-4mm}
    \caption{\rev{Visual comparison of our results to state-of-the-art approaches. Our recovered facial shapes are more faithful and recognizable. Our synthesized facial textures are with much higher quality, while there are undesired shadows/highlights or severe artifacts in the results of GANFIT and AvatarMe.}}
    \label{fig:comparisontosota}
\end{figure*}

\begin{figure*}[t]
    \centering
    \vspace{-3mm}
    \includegraphics[width=0.95\textwidth]{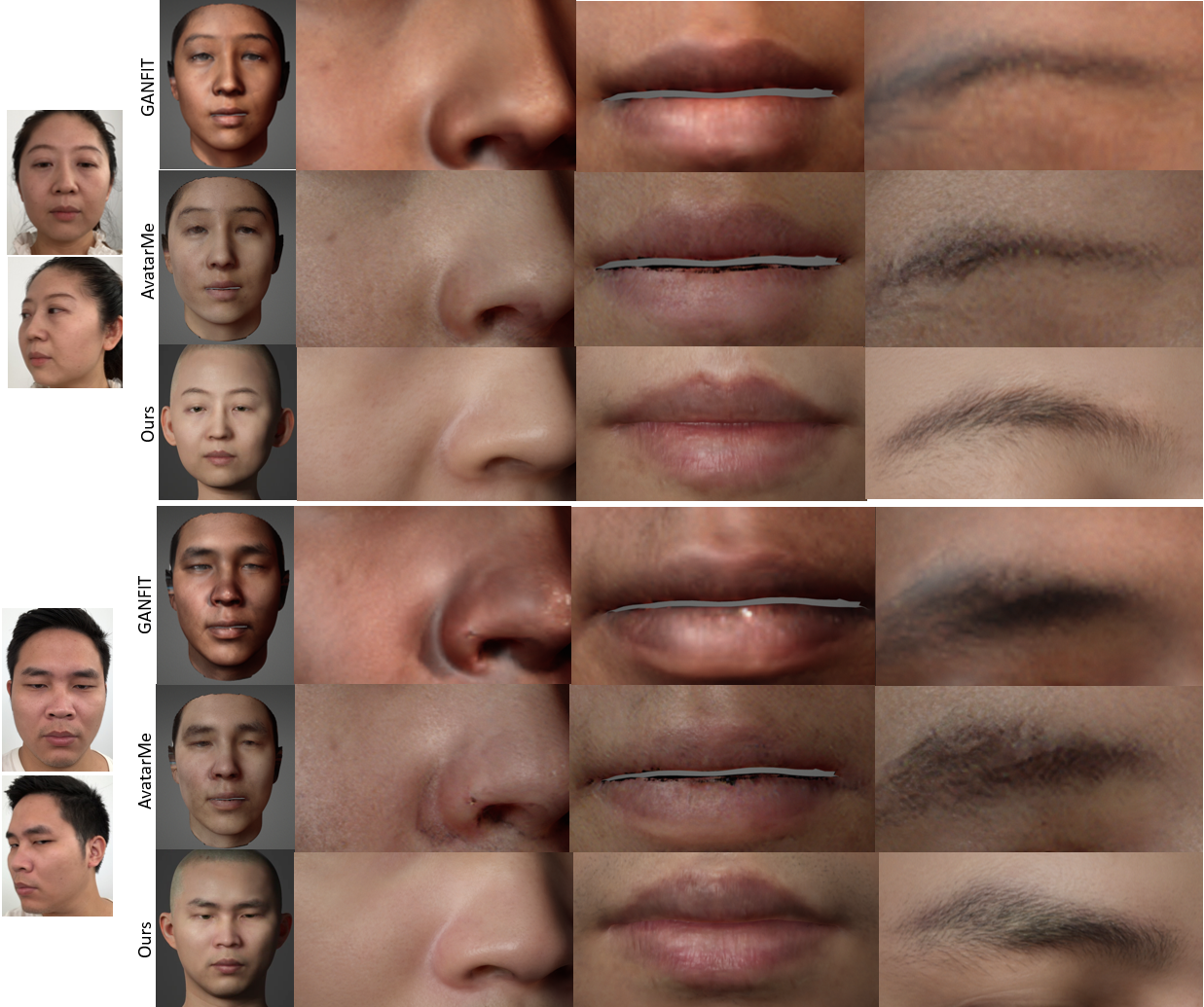}
    \vspace{-4mm}
    \caption{Rendered comparison. Our results are more faithful to input photos, with more realistic details. More results are in the supplementary document.}
    \label{fig:comparisontosotarendered}
\end{figure*}

\subsection{\rev{Comparison to State-of-the-art Approaches}}
\label{sec.compsota}

\rev{\subsubsection*{Visual Comparison}
We compare our results to two state-of-the-art 3D face reconstruction approaches, GANFIT \cite{gecer2019ganfit} and AvatarMe \cite{lattas2020avatarme}.
Fig. \ref{fig:comparisontosota} shows two examples of the results comparison. Fig. \ref{fig:comparisontosotarendered} shows the rendered comparison.
Our results are generally more faithful to input photos, with more accurate facial geometries and higher-quality texture maps.
Note that the inputs to GANFIT/AvatarMe are front-view single images, while our results are obtained with RGB-D selfie videos. }

\begin{table}[t]
    \centering
    \begin{tabular}{c|c}
    \hline
       \textbf{Method}  &  \textbf{Avg. ranking} \\
       \hline\hline
        AvatarMe \cite{lattas2020avatarme} & $7.0\%$ \\
        GANFIT \cite{gecer2019ganfit} & $5.1\%$ \\
        Ours & $\mathbf{4.5\%}$  \\
        \hline
    \end{tabular}
    \vspace{1mm}
    \caption{\rev{Identity verification using rendered images of the 3D reconstructions. The experiment is conducted with $36$ subjects' selfie data against more than $40,000$ distracting photos. Lower ranking means the rendered images are more recognizable to user photos from the point of view of a face recognition network. For example, the average ranking percentage $4.5\%$ means there are in average $4.5\%$ photos in the distracting dataset that are more similar to the rendered results than the users' own photos.}}
    \vspace{-5mm}
    \label{tab:id}
\end{table}

\subsubsection*{Identity Verification}
To further justify the recognizability of our results, we design a novel face verification experiment for numerical evaluation.
We collect selfie data from $36$ subjects and put their selfie photos into a large ``distracting'' face image dataset with over $40,000$ photos of Asian people.
Then the reconstructed results of the $36$ subjects are obtained using different approaches and rendered into realistic images using UE4.
We compute the feature distances of each rendered image to all the photos in the ``distracting'' dataset with a face recognition network \cite{deng2019arcface}.
Ideally, a user's selfie photo among all the ``distracting'' photos should be with the shortest distance to the rendered image.
Thus, for each rendered image, the distance ranking of a user's selfie photo among all the ``distracting'' photos is a good indicator of whether the rendered image is more recognizable to himself/herself or not.
The average ranking percentage of our identity verification experiment are shown in Table \ref{tab:id} (lower rankings are better).
Our results are generally more recognizable to user photos than AvatarMe and GANFIT.

\begin{figure}
    \centering
    \includegraphics[width=0.45\textwidth]{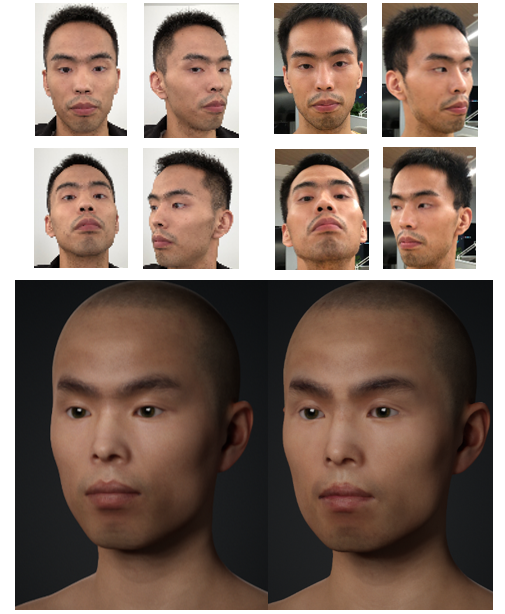}
    \vspace{-2mm}
    \caption{Results obtained using RGB-D data captured in different lighting conditions, but rendered in an identical lighting setting.
    The recovered facial structures are consistent in the two lighting conditions.
    Note that the skin color result in the right rendered image is a little bit more yellowish than the left.
    This is because the photos in the right is captured with a different white balance setting from the left.
    There is an inherent decoupling ambiguity between skin color and illumination in our approach.
    }
    \vspace{-3mm}
    \label{fig:robust_input}
\end{figure}

\begin{figure*}
    \centering
    \vspace{-1mm}
    \includegraphics[width=\textwidth]{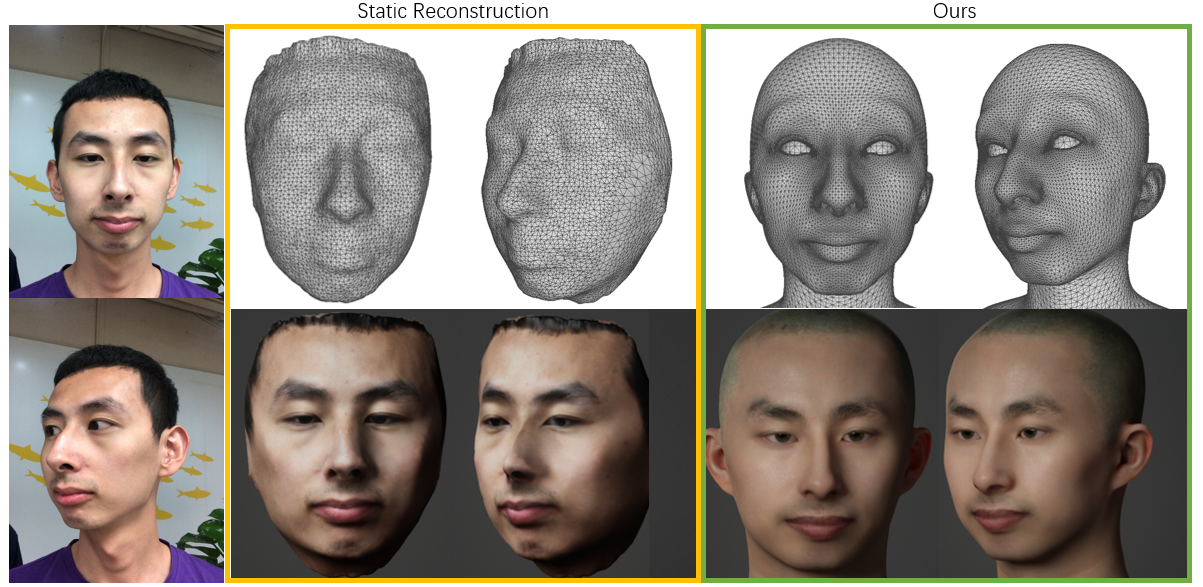}
    \vspace{-7mm}
    \caption{Comparison to model-free reconstruction (e.g., Bellus3D \shortcite{bellus3d2020}). The model-free reconstruction are not topologically consistent and are prone to flaws, which cause difficulties when being animated. The extracted textures contain undesired shadows and highlights that would result unnatural renderings. Our results are topological consistent and ready for animation. The high-quality albedo/normal maps make our rendering very realistic.}
    \vspace{-1mm}
    \label{fig:staticReconstruction}
\end{figure*}

\subsection{Comparison to Model-free Reconstruction}

We notice some commercial systems (e.g., Bellus3D \shortcite{bellus3d2020}) utilize RGB-D selfies to reconstruct static 3D face models and directly extract texture maps from input photos.
Their systems commonly employ a model-free reconstruction approach like KinectFusion \cite{newcombe2011kinectfusion} and the results usually seem very faithful to input photos.
However, there are several drawbacks in their results.
First, as shown in Fig. \ref{fig:staticReconstruction}, their reconstructed meshes are not topologically consistent and are prone to flaws.
It would be difficult to attach accessories and animate them.
Moreover, the extracted texture maps contain shadows and highlights, which are undesired since they cause severe unnatural issues when the rendered lighting is different from the captured lighting.
In comparison, our results are high-quality and ready for realistic rendering and animation (see Fig. \ref{fig:staticReconstruction}).

\subsection{Robustness to Different Inputs}

We conduct experiments for a user taking selfies in different lighting conditions (Fig. \ref{fig:robust_input}).
The recovered shape and reflectance remain consistent regardless of different lighting conditions and poses.
Note that there is an inherent decoupling ambiguity between skin color and illumination.
The resulting skin color in the right of the figure is actually a little bit more yellowish due to the yellower input photo.
However, the facial structures (like the eyebrow shape) in the resulting reflectance map are consistent.

\subsection{Rendered Results}

\begin{figure}
    \centering
    \vspace{-1mm}
    \includegraphics[width=0.5\textwidth]{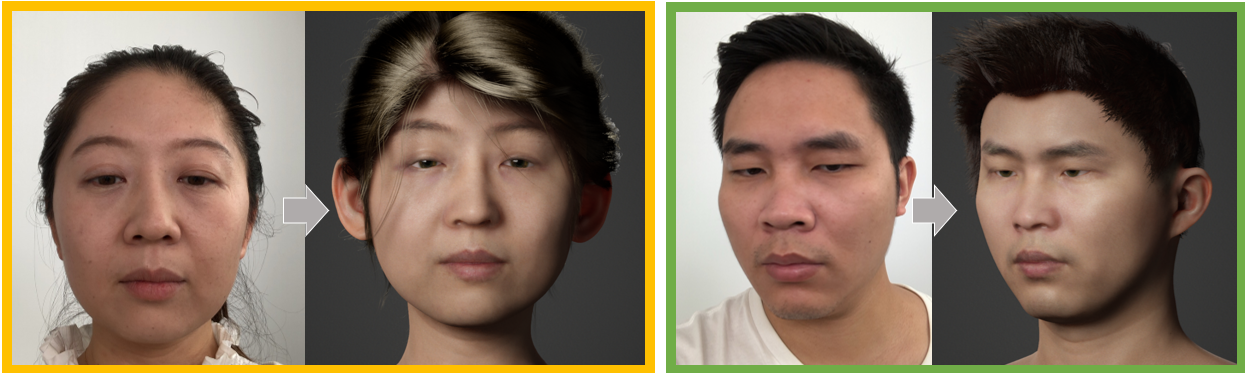}
    \vspace{-7mm}
    \caption{Rendering results with hair models.}
    \vspace{-3mm}
    \label{fig:different_hairtype}
\end{figure}

\begin{figure*}[t]
    \centering
    \includegraphics[width=0.995\textwidth]{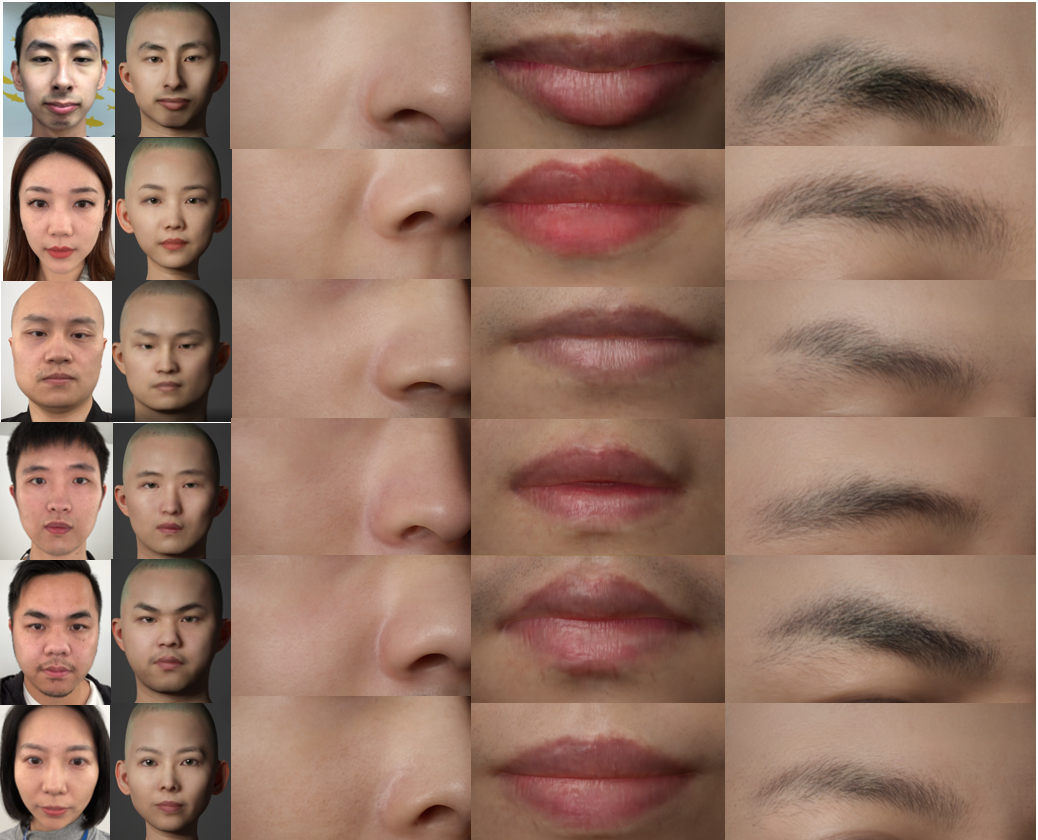}
    \vspace{-3mm}
    \caption{Rendered results with UE4 rendering engine. Our method can recover faithful face models with high-quality, realistic hair/pore/wrinkle details. Note that the selfie photos are with perspective camera projection, while the rendered results are with orthogonal camera projection.}
    \vspace{-3mm}
    \label{fig:render}
\end{figure*}

Fig. \ref{fig:render} shows some of our rendered results with UE4.
Thanks to the high-fidelity geometry and reflectance maps, the rendered results are realistic and faithful to input faces.
Note the wrinkle details on the lips, the pore-level details on the cheek, the hair-level details of the eyebrows.
Fig. \ref{fig:different_hairtype} shows two examples of our results with attached hair models, which are retrieved from our hair model database by performing hairstyle classification on the selfie photos.
More results are in the supplementary video.

\begin{figure}
    \centering
    \includegraphics[width=0.5\textwidth]{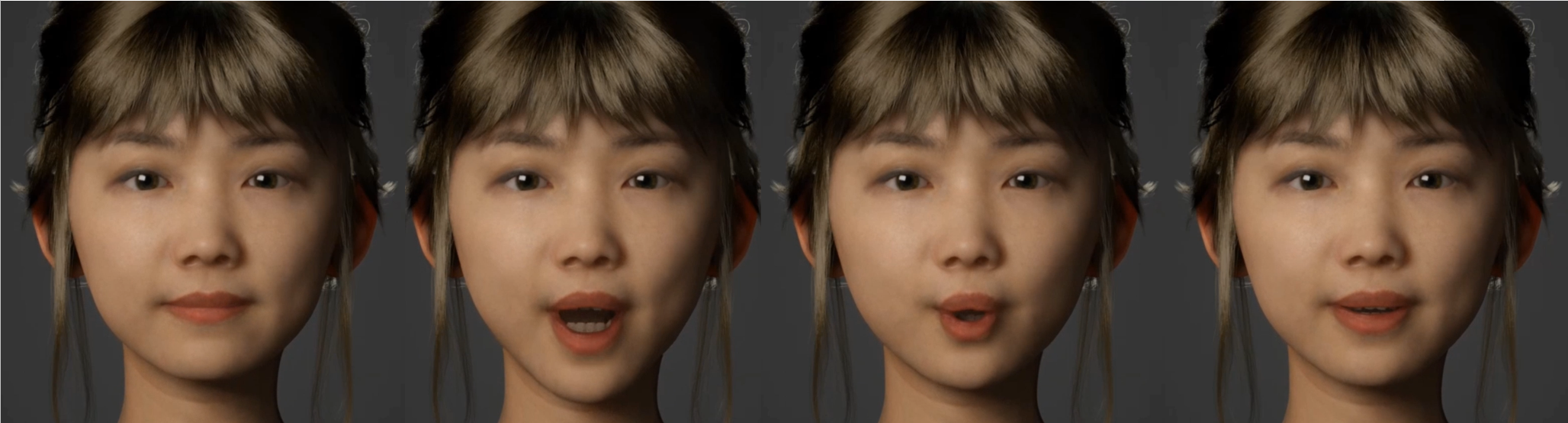}
    \vspace{-7mm}
    \caption{Snapshots of our lip-sync animation. See supplementary video.}
    \vspace{-3mm}
    \label{fig:lip_sync_animation}
\end{figure}

\begin{figure}
    \centering
    \vspace{-1mm}
    \includegraphics[width=0.5\textwidth]{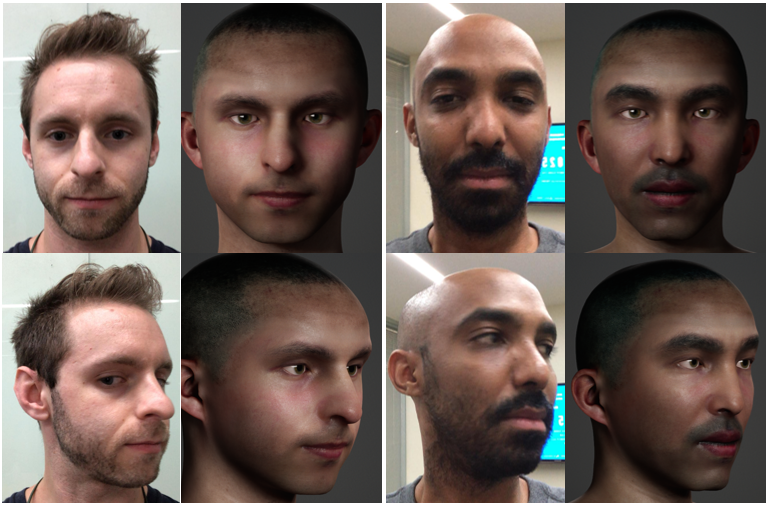}
    \vspace{-7mm}
    \caption{Results on other ethnicities different from our 3DMM data source. The results roughly resemble the subjects but lack ethnicity-specific features.}
    \vspace{-3mm}
    \label{fig:failurecases}
\end{figure}

\subsection{Limitations}

Our approach does not take into account cross-ethnicity generalization.
Since our 3DMM is constructed from East-Asian subjects, our evaluations are designed on the same ethnical population.
We tried to directly apply our approach on some people from other ethnicities and Fig. \ref{fig:failurecases} shows two examples.
Although the results roughly resemble the subjects being captured, some ethnicity-specific facial features are not recovered.
Besides, our approach cannot model facial hairs like moustache and beard.
Children or aged people beyond the age scope of our dataset are also not considered in our approach.

\rev{Another limitation of our approach is that less common details like moles or freckles are not captured. The problem needs to be tackled by capturing these details and then sending them into the detail synthesis networks, which would be left as future work. Besides, the skin tone variation in the reconstructed albedo is also limited, due to the inherent decoupling ambiguity between skin color and illumination (see Figs. \ref{fig:robust_input} and \ref{fig:failurecases} for examples). It can be solved with a color checker while taking selfies.}

%%%%%%%%%%%%%%%%%%%%%%%%%%%%%%%%%%
%%%%%%%%%%%%%%%%%%%%%%%%%%%%%%%%%%
\section{Applications}
%TTS driven results
The avatars created with our method are animation-ready.
Animations can be retargeted from existing characters \cite{bouaziz2014semi}, or interactively keyframe-posed \cite{ichim2015dynamic}, or even transferred from facial tracking applications \cite{weise2011realtime}.
We demonstrate an application of lip-sync animation in the supplementary video, where a real-time multimodal synthesis system  \cite{yu2019durian} is adopted to simultaneously synthesize speech and expression blendshape weights given input texts.
Fig. \ref{fig:lip_sync_animation} shows several snapshots of the animation.
The application enables users to conveniently create high-fidelity, realistic digital humans that can be interacted with in real time.
We also include another lip-sync animation result driven by speech inputs \cite{huang2020speaker} in the supplementary video.

%%%%%%%%%%%%%%%%%%%%%%%%%%%%%%%%%%
%%%%%%%%%%%%%%%%%%%%%%%%%%%%%%%%%%
\section{Conclusion}

We have introduced a fully automatic system that can produce high-fidelity 3D facial avatars with a commercial RGB-D selfie camera.
The system is robust, efficient, and consumer-friendly.
The total acquisition and processing for a user can be finished in less than 30 seconds.
The generated geometry models and reflectance maps are in very high fidelity and quality.
With a physically based renderer, the assets can be used to render highly realistic digital humans.
Our system provides an excellent consumer-level solution for users to create high-fidelity digital humans.

\textbf{Future Work~~} The animation with generic expression blendshapes are not satisfactory. We intend to extend our system to capture personalized expression blendshapes like Ichim et al. \shortcite{ichim2015dynamic}.
Besides, the current system employs very simple approaches to handle accessories like hair, eyeballs, and teeth.
We intend to incorporate more advanced methods to model accessories.

%%%%%%%%%%%%%%%%%%%%%%%%%%%%%%%%%%
%%%%%%%%%%%%%%%%%%%%%%%%%%%%%%%%%%
% DO NOT INCLUDE ACKNOWLEDGMENTS IN AN ANONYMOUS SUBMISSION TO SIGGRAPH 2019
\begin{acks}

We would like to thank Cheng Ge and other colleagues at Tencent NExT Studios for valuable discussions; Shaobing Zhang, Han Liu, Caisheng Ouyang, Yanfeng Zhang, and other colleagues at Tencent AI Lab for helping us with the videos; and all the subjects for allowing us to use their selfie data for testing.

\end{acks}

% Bibliography
\bibliographystyle{ACM-Reference-Format}
\bibliography{3dfacebib}

% % Appendix
% \appendix
% \section{Switching Times}

% In this appendix, we measure the channel switching time of Micaz
% \cite{CROSSBOW} sensor devices.  In our experiments, one mote
% alternatingly switches between Channels~11 and~12. Every time after
% the node switches to a channel, it sends out a packet immediately and
% then changes to a new channel as soon as the transmission is finished.
% We measure the number of packets the test mote can send in 10 seconds,
% denoted as $N_{1}$. In contrast, we also measure the same value of the
% test mote without switching channels, denoted as $N_{2}$. We calculate
% the channel-switching time $s$ as
% \begin{displaymath}%
% s=\frac{10}{N_{1}}-\frac{10}{N_{2}}.
% \end{displaymath}%
% By repeating the experiments 100 times, we get the average
% channel-switching time of Micaz motes: 24.3\,$\mu$s.

\end{document}